\documentclass[sigconf]{acmart}
\usepackage{graphicx}
\usepackage{subfigure}
\usepackage{amsmath}
\usepackage{enumitem}
\usepackage{balance}

\AtBeginDocument{%
  }

\setcopyright{acmlicensed}
\copyrightyear{2024} 
\acmYear{2024} 
\setcopyright{acmlicensed}\acmConference[KDD '24]{Proceedings of the 30th
 ACM SIGKDD Conference on Knowledge Discovery and Data Mining}{August
 25--29, 2024}{Barcelona, Spain}
 \acmBooktitle{Proceedings of the 30th ACM SIGKDD Conference on Knowledge
 Discovery and Data Mining (KDD '24), August 25--29, 2024, Barcelona, Spain}
 \acmDOI{10.1145/3637528.3671936}
 \acmISBN{979-8-4007-0490-1/24/08}

\begin{document}


\title{Cluster-Wide Task Slowdown Detection in Cloud System}
% \title{Cluster-Wide Task Slowdown Detection in Cloud Systems}

\author{Feiyi Chen}
\email{chenfeiyi@zju.edu.cn}
\affiliation{%
  \institution{Zhejiang University, Alibaba Group}
  \city{Hangzhou}
  \country{China}
}
\author{Yingying Zhang}
\email{congrong.zyy@alibaba-inc.com}
\affiliation{%
  \institution{ Alibaba Group}
  \city{Hangzhou}
  \country{China}
}
\author{Lunting Fan}
\email{lunting.fan@taobao.com}
\affiliation{%
  \institution{ Alibaba Group}
  \city{Hangzhou}
  \country{China}
}
\author{Yuxuan Liang}
\email{yuxliang@outlook.com}
\affiliation{%
  \institution{The Hong Kong University of Science and Technology (Guangzhou)}
  \city{Guangzhou}
  \country{China}
}
\author{Guansong Pang}
\email{gspang@smu.edu.sg}
\affiliation{%
  \institution{Singapore Management University}
  \city{Singapore}
  \country{Singapore}
}
\author{Qingsong Wen}
\email{qingsongedu@gmail.com}
\affiliation{%
  \institution{Squirrel AI}
  \country{Bellevue, USA}
}
\author{Shuiguang Deng}
\authornote{*Corresponding authors}
\email{dengsg@zju.edu.cn}
\affiliation{%
  \institution{Zhejiang University}
  \city{Hangzhou}
  \country{China}
}

\renewcommand{\shortauthors}{Feiyi Chen et al.}

\begin{abstract}
    Slow task detection is a critical problem in cloud operation and maintenance since it is highly related to user experience and can bring substantial liquidated damages. Most anomaly detection methods detect it from a single-task aspect. However, considering millions of concurrent tasks in large-scale cloud computing clusters, it becomes impractical and inefficient. Moreover, single-task slowdowns are very common and do not necessarily indicate a malfunction of a cluster due to its violent fluctuation nature in a virtual environment. Thus, we shift our attention to cluster-wide task slowdowns by utilizing the duration time distribution of tasks across a cluster, so that the computation complexity is not relevant to the number of tasks.
    The task duration time distribution often exhibits compound periodicity and local exceptional fluctuations over time. Though transformer-based methods are one of the most powerful methods to capture these time series normal variation patterns, we empirically find and theoretically explain the flaw of the standard attention mechanism in reconstructing subperiods with low amplitude when dealing with compound periodicity. 
  To tackle these challenges, we propose SORN (i.e., \underline{S}kimming \underline{O}ff subperiods in descending amplitude order and \underline{R}econstructing \underline{N}on-slowing fluctuation), which consists of a Skimming Attention mechanism to reconstruct the compound periodicity and a Neural Optimal Transport module to distinguish cluster-wide slowdowns from other exceptional fluctuations. Furthermore, since anomalies in the training set are inevitable in a practical scenario, we propose a picky loss function, which adaptively assigns higher weights to reliable time slots in the training set. Extensive experiments demonstrate that SORN outperforms state-of-the-art methods on multiple real-world industrial datasets.
\end{abstract}



\maketitle

\section{Introduction}

\begin{figure*}[t]
  \centering 
  \subfigure[The task duration time distribution]{
    \includegraphics[width=0.3\linewidth]{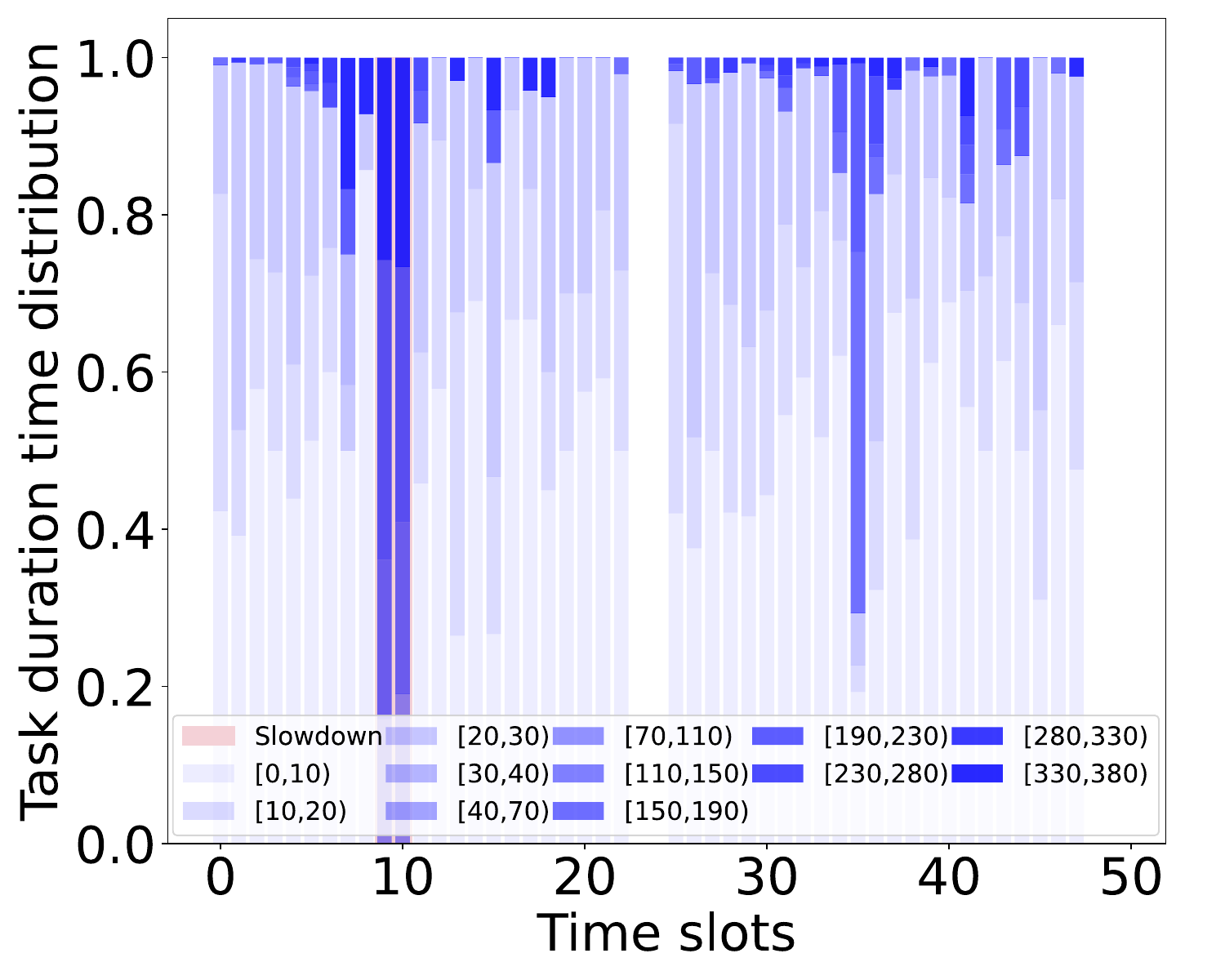}
    \label{fig:distribution}
  }
  \hfill
  \subfigure[The compound periodicity]{
      \includegraphics[width=0.3\linewidth]{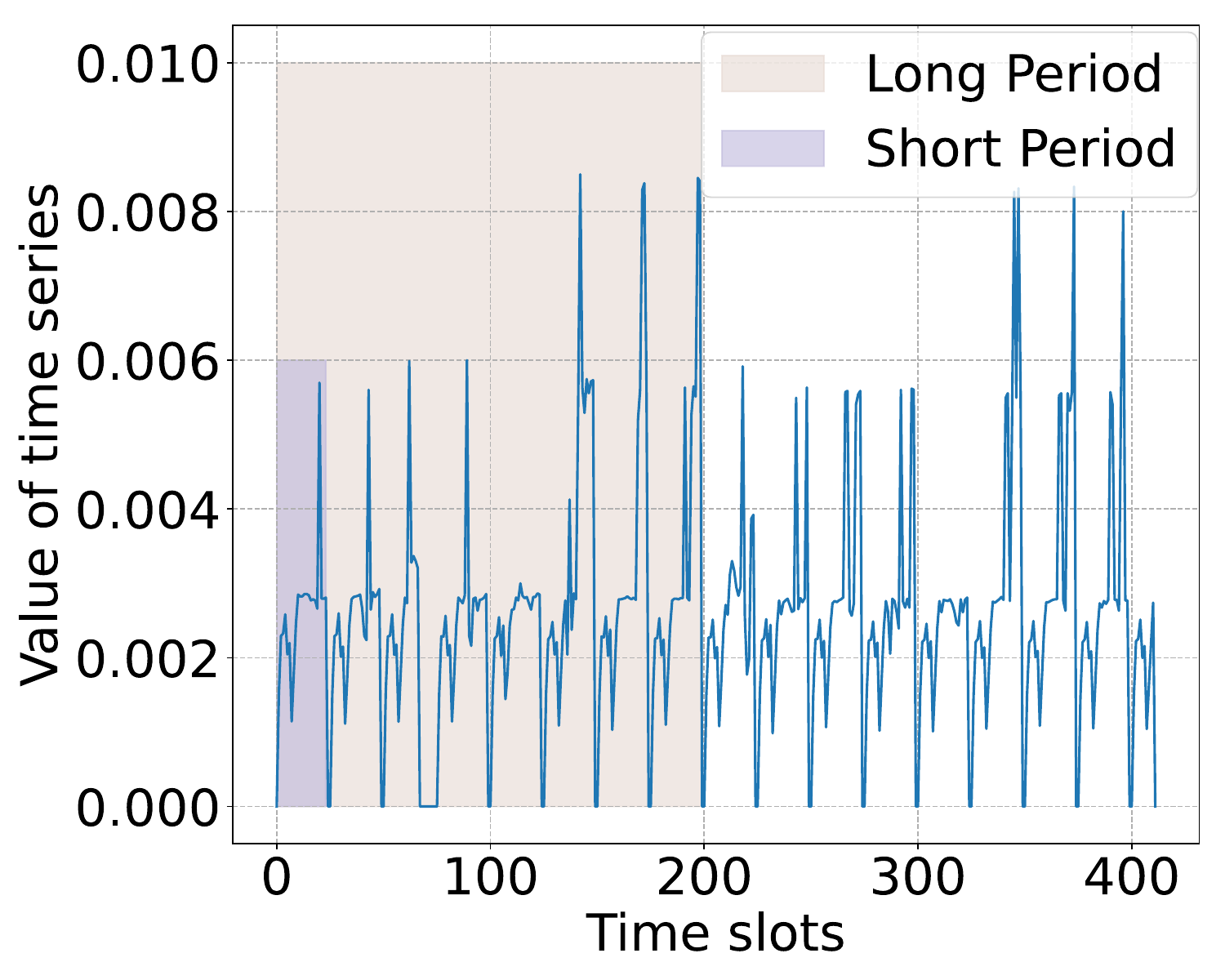}
      \label{fig:compound}
  }
  \hfill
  \subfigure[The reconstruction series of attention]{
    \includegraphics[width=0.3\linewidth]{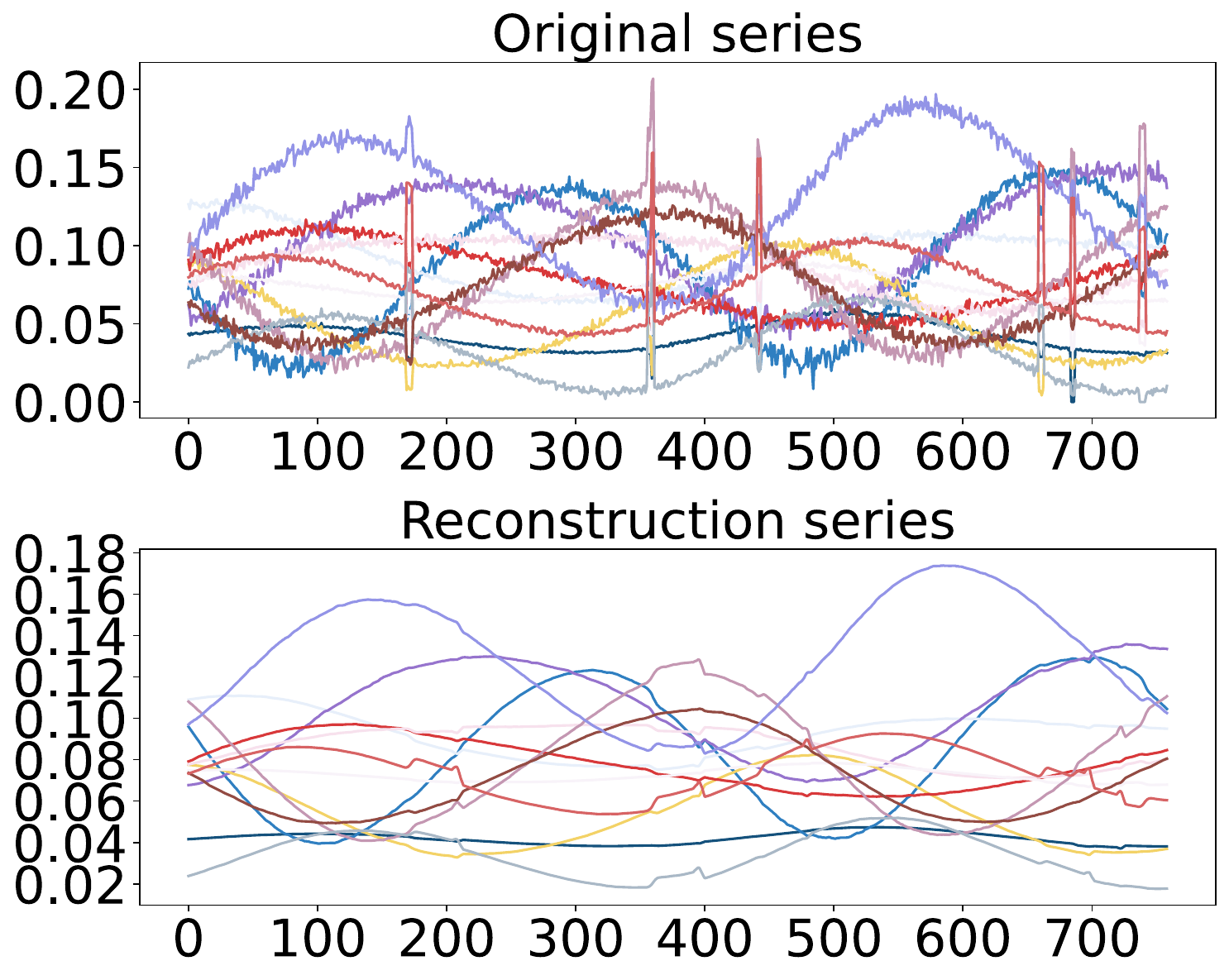}
    \label{fig:reconstruction}
    }\vspace{-3mm}
  \caption{(a) At each time slot, we use a stacked histogram bar to plot the frequency distribution of the duration time at that slot. We use a darker color to denote the interval requiring more duration time. The stacked histogram bar is ordered in time order. (b) The compound periodicity of task duration time. (c) The original series and series reconstructed by standard attention are plotted in one figure, where the subperiod with low amplitude can not be well reconstructed.}
\end{figure*}

Slow task detection is a critical issue in cloud operations and maintenance, as it directly impacts user experience and can lead to significant penalties for service level agreement violations \cite{upadhyay2020stdads}. Most existing anomaly detection methods focus on detecting task slowdowns at the individual task level \cite{ma2021jump, yang2023dcdetector,zhang2019deep,su2019robust}. However, with millions of tasks running concurrently \cite{ma2021jump,zhang2021cloudrca} in large-scale cloud computing clusters, these approaches become impractical and inefficient. Moreover, single-task slowdowns are common and may not indicate a cluster malfunction, given the random and dramatic fluctuations in task duration time within a virtual environment.
To address these challenges, we pivot towards detecting slowdowns on a cluster-wide scale, which are more indicative of cluster malfunctions and can be identified without examining each individual task. Furthermore, unlike the random fluctuations observed in single-task duration time, the duration time of cluster-wide tasks exhibits more regular patterns, making slowdown detection more feasible. Particularly, we detect cluster-wide task slowdowns using the duration time distribution of a cluster, as illustrated in Fig.~\ref{fig:distribution}, in which for each time slot we partition the range of task duration time into intervals and calculate the proportion of tasks falling into each interval.
This strategic shift not only significantly reduces the computational complexity of our algorithm, making it independent of the number of tasks, but also enhances the accuracy of cluster malfunction detection.

Nonetheless, the distribution of normal task duration time is not stable but varies over time. Hence, there arises a necessity to discern the patterns of distribution variation and differentiate routine slowdowns from anomalies. Among the various methods for extracting normal patterns, transformer-based methods stand out as one of the most powerful and effective unsupervised anomaly detection approaches, resulting in numerous distinguished methods \cite{li2023prototype, DBLP:conf/iclr/XuWWL22, tuli2022tranad,yang2023dcdetector}.
Despite the abundance of powerful neural networks available for normal pattern extraction, several challenges persist:
\begin{itemize}[leftmargin=*]
  \item \emph{Compound periodicity}: The distribution of cluster-wide task duration time often exhibits compound periodic variation patterns. Since different tasks exhibit different periodicity, the periodicity of cluster-wide task duration time distribution is compound and complicated. For example, in Fig. \ref{fig:compound}, it shows periodicity on both a weekly and daily basis. As depicted in Fig.\ref{fig:reconstruction}, when integrating two periodicities with different amplitudes and frequencies into a unified representation, the attention mechanism shows subpar performance in reconstructing the subperiodicity with relatively low amplitude in the presence of compound periodicity.
  
  \item \emph{Non-slowing exceptional fluctuations}: 
The temporal evolution of task duration time within the cluster manifests periodic characteristics on a global scale, interspersing with localized non-periodic exceptional fluctuations. Within these exceptional fluctuations, only a small fraction corresponds to cluster-wide slowdowns, while others are not the focus of our work (e.g., we are not concerned about exceptional task speedups). However, traditional anomaly detection methods can not reconstruct all of the exceptional fluctuations well and detect all of them as anomalies.
To distinguish cluster-wide task slowdowns, it is imperative to accurately reconstruct other exceptional fluctuations while excluding the cluster-wide slowdowns.
  
  \item \emph{Anomalies in the training set}: 
In consideration of the substantial costs linked to manually labeling anomalies, our methodology has been intentionally crafted to function in an unsupervised manner.  Nevertheless, it is noteworthy that several unsupervised methods operate on the assumption that anomalies are infrequent within the training set, a premise that tends to be overly optimistic in practical scenarios.
\end{itemize}

Addressing these challenges is imperative for improving the detection accuracy of compound periodic time series and enhancing model robustness against anomaly contamination in the training set. Therefore, we propose SORN, which \underline{S}kims \underline{O}ff the subperiodicity with different amplitudes layer by layer and selectively \underline{R}econstructs the \underline{N}on-slowing fluctuations excluding the cluster-wide task slowdowns. It contains three innovative mechanisms to tackle the aforementioned three issues correspondingly: Skimming Attention, Neural Optimal Transport (OT), and Picky Loss. 

Specifically, we first theoretically prove that the standard attention mechanism tends to allocate more attention to subperiods with higher amplitudes in compound periodic time series. This bias prevents it from effectively reconstructing subperiods with relatively low amplitudes. Building on this analysis, we introduce a skimming attention mechanism to capture the compound periodicity pattern, where we sequentially skim off subperiods from the original sequence in descending order of amplitudes and reconstruct iteratively from the remaining series. In this way, the subperiods with higher amplitudes are initially well reconstructed and skimmed off from the original time series. After that, the subperiods with low amplitudes in the original series become subperiods with relatively high amplitudes in the remaining series and can be better reconstructed.

Subsequently, we use a Neural OT module to adjust the reconstructed series of skimming attention, where we innovatively transform the traditional optimal transport problem into a neural network, and by intricately designing a transportation cost matrix, we can selectively reconstruct the non-slowing fluctuations.

Furthermore, to mitigate the negative effect of anomaly contamination in the training set, we design a novel picky loss function, which allocates different weights to time slots in the loss function according to their reliability.

Accordingly, this work presents several novel and distinctive contributions to the field of cluster-wide slow task detection:
\begin{itemize}[leftmargin=*]
  \item We present the first attempt to formalize the cluster-wide slowdown problem with the identification of the problem specifications and relevant challenges.
  \item We provide a theoretical explanation for the limitations of the standard attention mechanism in effectively reconstructing subperiods with low amplitude in compound periodicity. Moreover, we introduce a novel skimming attention mechanism designed to extract subperiodic components with varying amplitudes and aggregate them to ensure accurate reconstruction of both high and low-amplitude subperiods.
  \item We introduce a novel Neural OT module tailored to reconstruct the normal non-periodic fluctuations observed in the duration time distribution, while effectively filtering out the cluster-wide slow-down anomalies.
  \item We propose a picky loss function that assigns higher weights to reliable time slots within the loss function.
\end{itemize}
Besides, we conducted extensive experiments and demonstrated that our method outperforms the state-of-the-art (SOTA) methods in F1 score on real-world industrial datasets.

\section{Preliminary}
\subsection{Optimal Transport (OT)}
It is given a set of value intervals $I=\{(s_1,s_2],(s_2,s_3],\dots,(s_{n-1},s_n]\}$ and two distributions $\mathbf{a}\in R^N$ and $\mathbf{b} \in R^N$, where $\mathbf{a}_i=P(s_i<x\leq s_{i+1}), x\sim \mathbf{a}$. 
Similarly, $\mathbf{b}_i=P(s_i<x\leq s_{i+1}),x\sim \mathbf{b}$.
The Optimal Transport problem aims at transforming distribution $\mathbf{a}$ to $\mathbf{b}$ by moving a fraction of the amount in each interval of $\mathbf{a}$ to another interval. Moving a unit from $j^{th}$ interval to $i^{th}$ interval costs a price $C_{i,j}$. The Optimal Transport problem gropes for an optimal transport strategy $P$ costing the lowest price, where $P_{i,j}$ denotes the amount of unit moving from $j^{th}$ interval to the $i^{th}$, as shown in Eq.\ref{Eq:OT}, where $<P,C>$ denotes the Frobenius dot-product.
\begin{equation}
    \label{Eq:OT}
    \begin{split}
    \min_{P}<P&,C>, \\
    s.t.\ P \cdot \vec{1}=\mathbf{b},& \ P^T \cdot \vec{1}=\mathbf{a}.
    \end{split}
\end{equation}

\subsection{Problem Setup}
\textbf{Definition 1.} $f_t$ and $f_{t}^*$ are used to denote the real-time distribution and expected distribution at time slot $t$. $f_t(\alpha)$ and $f_t^*(\alpha)$ are used to denote the $\alpha$-quantile of distribution $f_t$ and $f_t^*$. $\mathcal{T}$ is used to denote the threshold for tolerable fluctuation range of duration time distribution.
\\
\textbf{Definition 2.} If there is a slowdown at time slot $t$, then $\max_{\alpha} f_t(\alpha)-f^*_t(\alpha) > \mathcal{T}, \forall \alpha$.
\\
\textbf{Definition 3.} (Input data \& output data) Given a set of intervals $I=\{[s_1,s_2),[s_2,s_3),\dots,[s_{D},s_{D+1})\}$, the input data is a $T-$length and $D$ dimensional multivariate time series $x\in R^{T\times D}$, where $x[t,d]$ is the number of tasks whose duration time falls into the $d^{th}$ interval $[s_d,s_{d+1})$. The reconstruction series of SORN is denoted by $\dot{\tilde{x}}$.
\\
\textbf{Problem Formalization.} We use a SORN to obtain a reconstruction series $\dot{\tilde{x}}$ from the original input data $x$. Subsequently, we use an anomaly score function $\operatorname{AnomalyScore}(\dot{\tilde{x}},x,I)$. We aim to maximize the anomaly score gap between the slow-down time slots and the others.

\section{Methodology}
\begin{figure*}[t]
    \centering 
    \subfigure[The model architecture of SORN]{
        \includegraphics[width=0.47\linewidth]{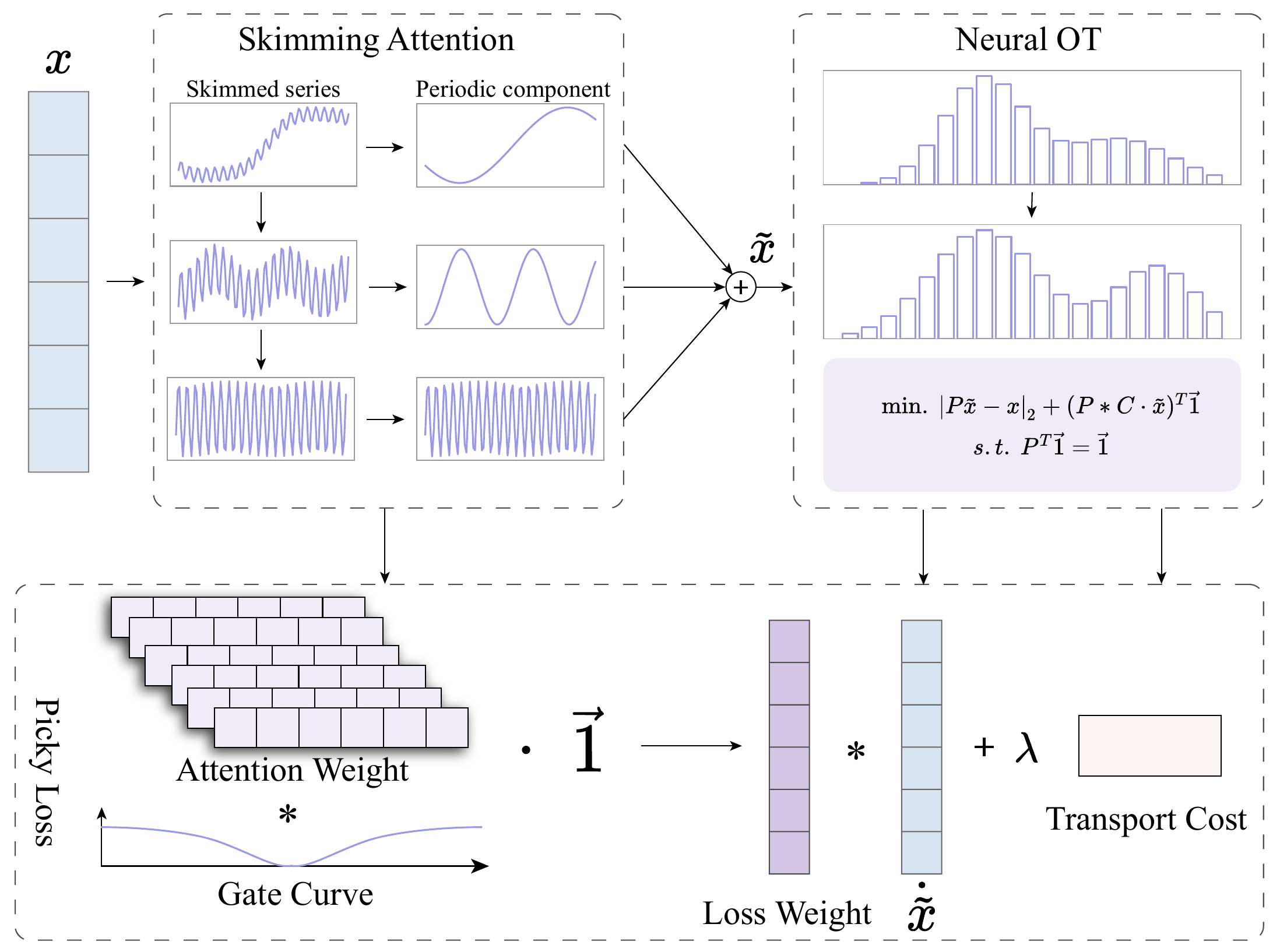}
        \label{fig:modelArch}
    }
    \hfill
    \subfigure[A layer of Skimming Attention]{
    \includegraphics[width=0.48\linewidth]{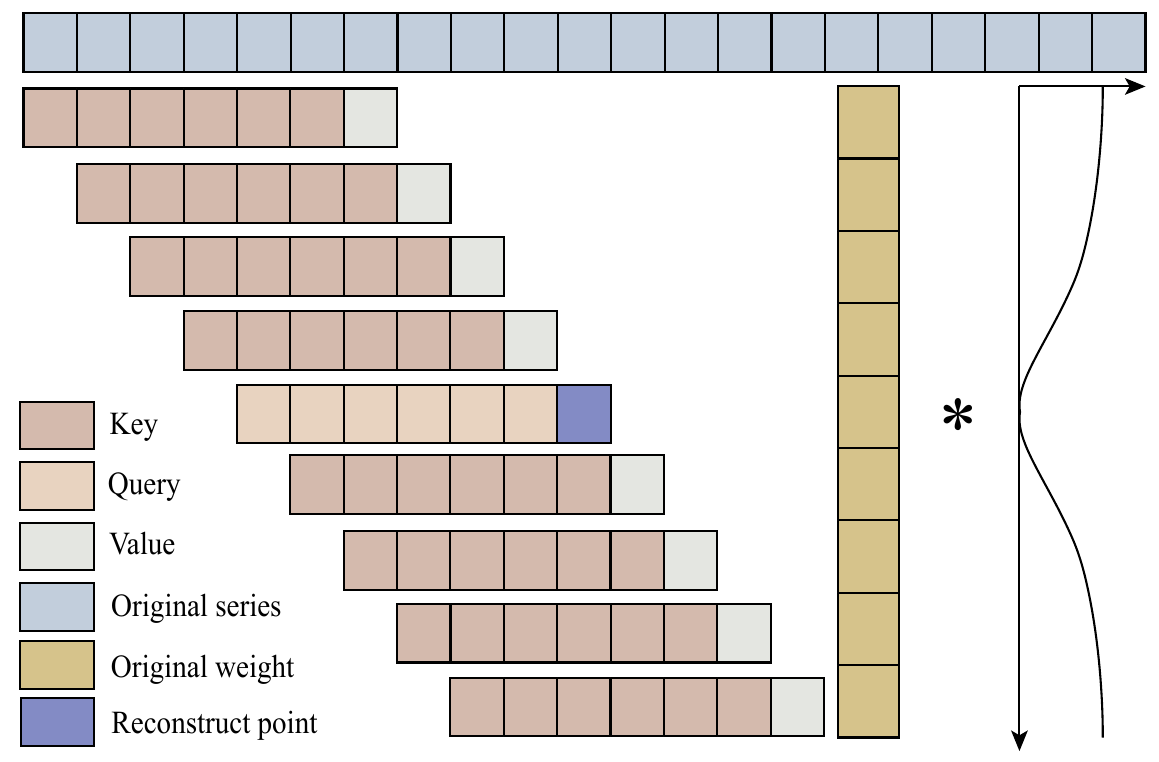}
    \label{fig:skimmingAtten}
    }\vspace{-3mm}
    \caption{The model architecture of the proposed SORN algorithm.}
\end{figure*}

The overview of SORN is depicted in Fig. \ref{fig:modelArch}. We sequentially mask each time slot in $x$ and employ a multi-layer Skimming Attention mechanism to reconstruct the time slot by leveraging compound periodic information. Subsequently, we utilize Neural OT to fine-tune the reconstructed series obtained from Skimming Attention, capturing aperiodic but typical fluctuations in the time series.
Finally, we apply the picky loss function to assign higher weights to normal time slots while assigning lower weights to occasional anomalous slots in the loss function.

\begin{figure*}
  \centering 
  \subfigure[The amplitude of different periods]{
  \includegraphics[width=0.23\linewidth]{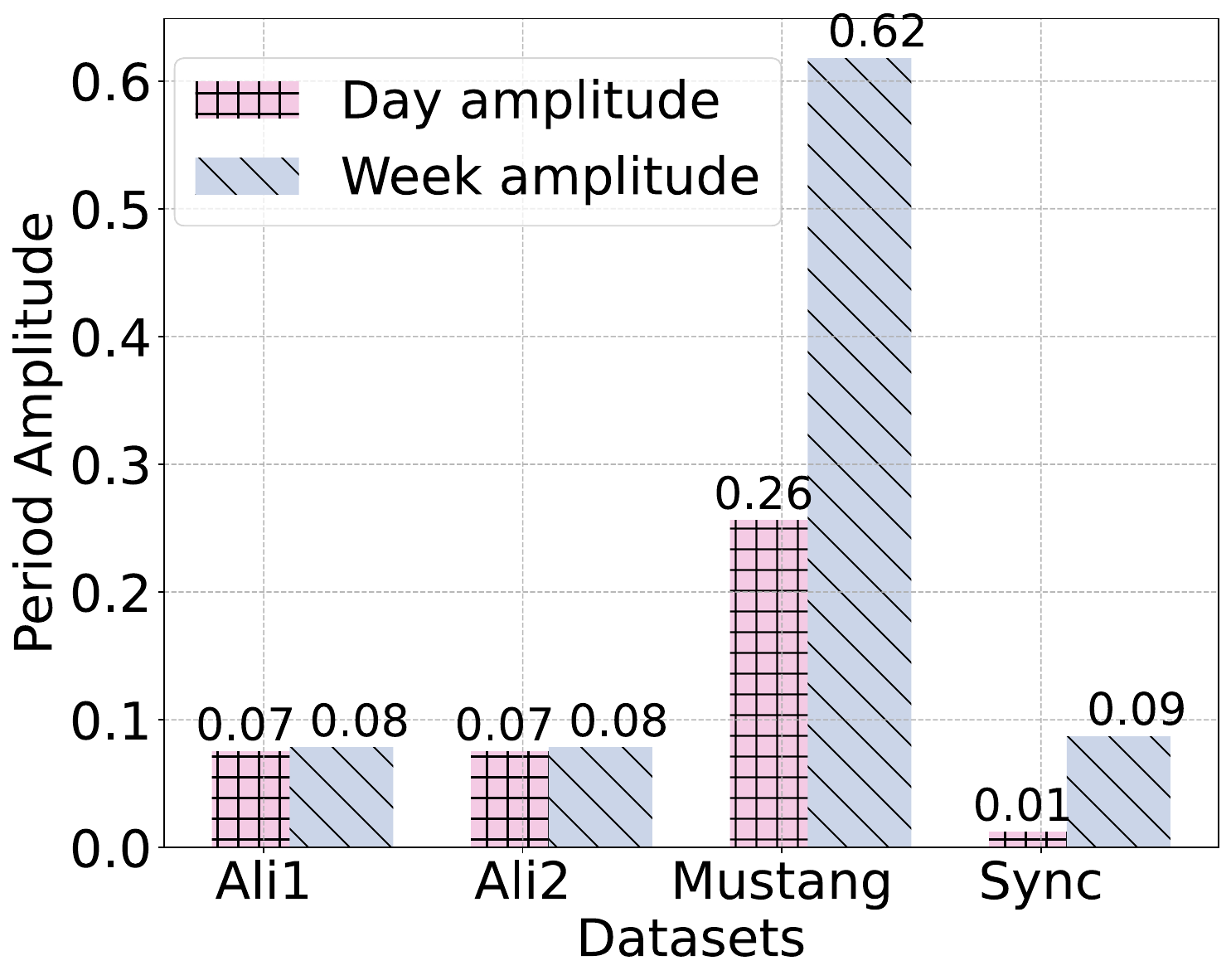}
  \label{fig:amplitude}
  }
  \hfill 
  \subfigure[Attention weights along different subperiods]{
  \includegraphics[width=0.24\linewidth]{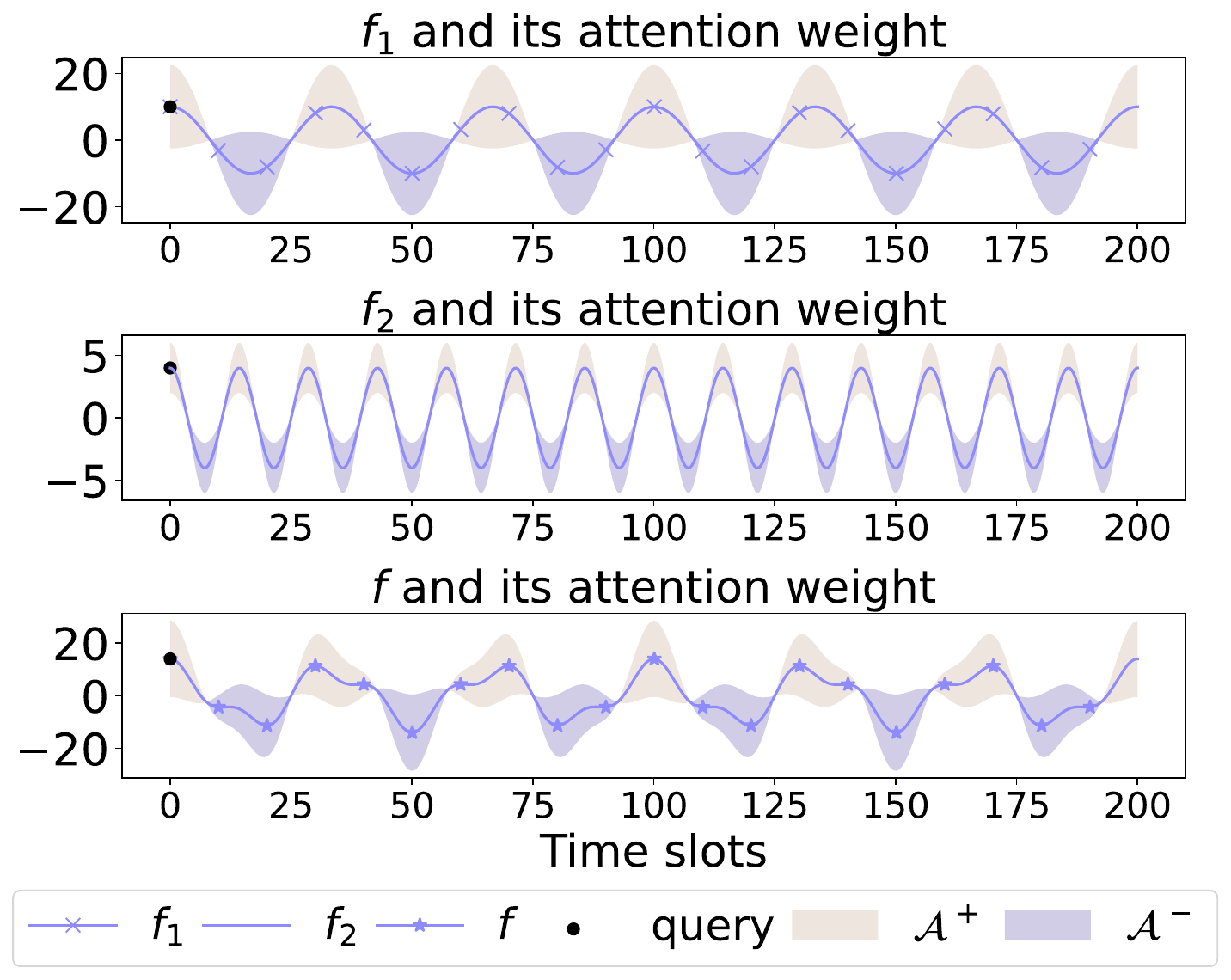}
  \label{fig:attention}
  }
  \hfill
  \subfigure[The series from different Skimming Attention layers]{
  \includegraphics[width=0.21\linewidth]{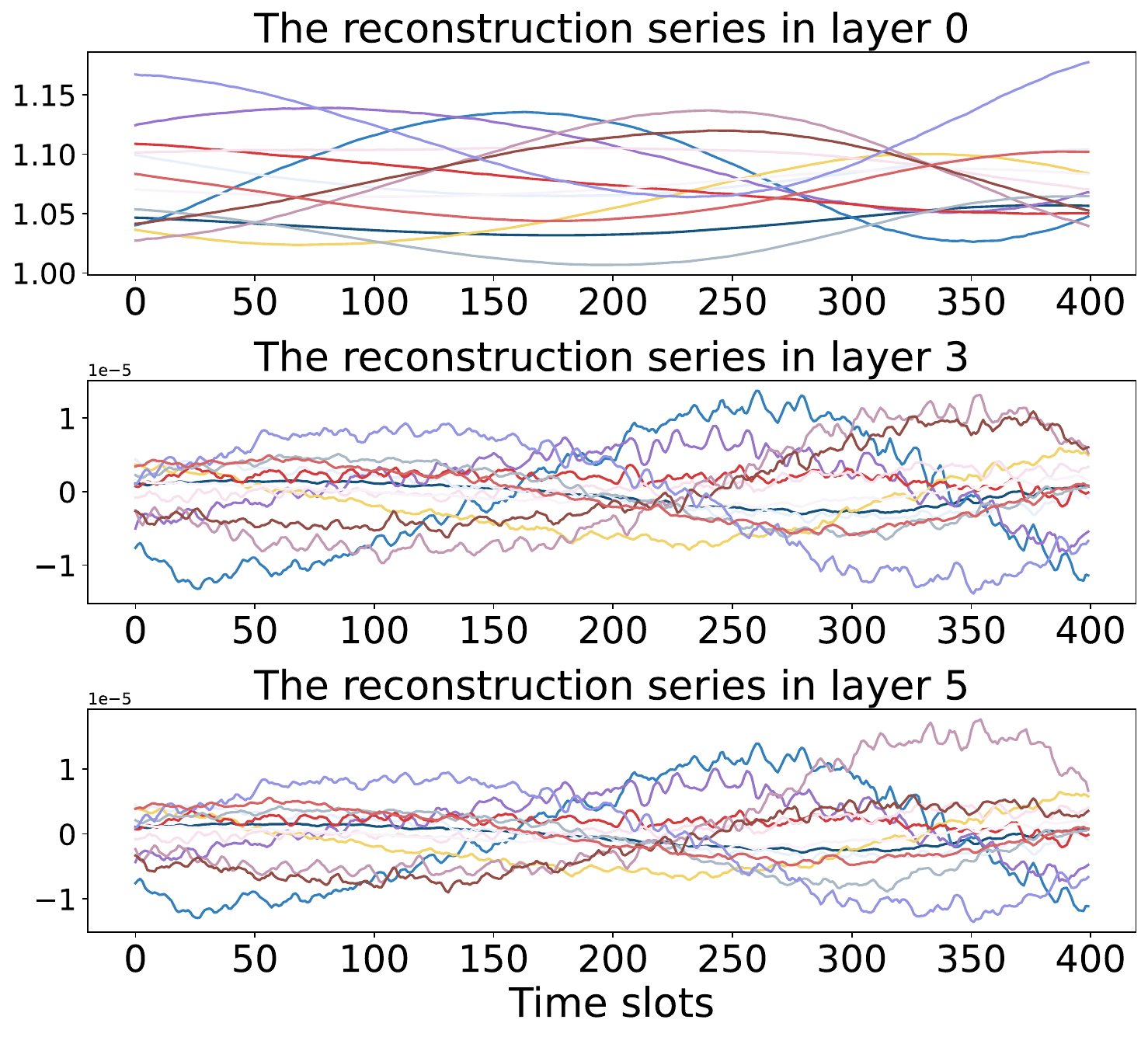}
  \label{fig:skimSeries}
  }
  \hfill
  \subfigure[The reconstruction and loss weight of SORN]{
      \includegraphics[width=0.21\linewidth]{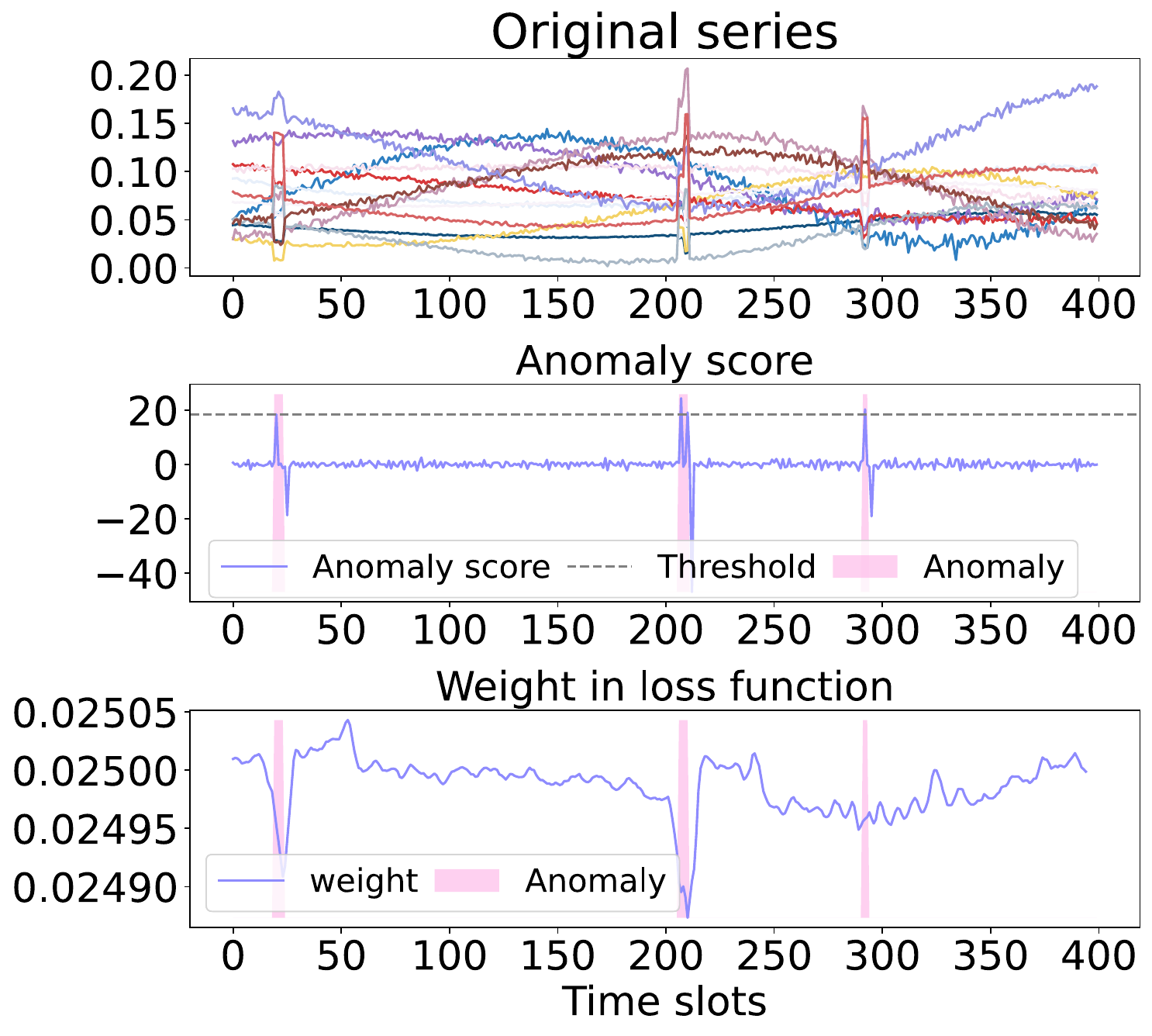}
      \label{fig:visualize}
  }
  \hfill \vspace{-3mm}
  \caption{ (a) The figure shows different amplitudes of different subperiods;  (b) The figure shows attention weight along different subperiods in $f(t)$. The width of the shadow is the value of the attention weight divided by 100 at the corresponding time slot. To distinguish the positive attention weight and negative attention weight we plot them in different colors and denote them by $\mathcal{A}^+$ and $\mathcal{A}^-$ respectively. (c) \& (d) The visualization of SORN.}
\end{figure*}

\subsection{Skimming Attention}
\label{sec:skimmingAtt}
The duration time distribution usually exhibits compound periodic fluctuations, as shown in Fig.~\ref{fig:compound}. In a compound periodic series, different subperiods usually have different amplitudes (i.e., variation range)~\cite{wen2020fast}, as shown in Fig.~\ref{fig:amplitude}. When dealing with this kind of compound periodicity, the standard attention mechanism falls short in reconstructing the subperiod with low amplitude, as shown in Fig.\ref{fig:reconstruction}, where we fuse two periodicities with different amplitudes and frequency, the standard attention only reconstructs the one with higher amplitude well.
We theoretically explain this phenomenon in Theorem 1 and Theorem 2, where we prove that a self-attention mechanism pays more attention to the subperiod with relatively higher amplitude in compound periodic series, which degrades the performance of reconstructing the subperiods with lower amplitudes in compound periodic series. 
Thus, we propose a skimming attention that masks each time slot alternatively and aims at reconstructing it by compound periodic information. There are two challenges to achieving this. On the one hand, we need to prevent it from reconstructing time slots only by leveraging the similarity of adjacent time slots in each layer but neglecting the periodic information. On the other hand, we need to reconstruct every subperiod well rather than just those with high amplitudes.

We deduce Theorems 1-2 using the same setting as the self-attention mechanism in a patching transformer \cite{DBLP:conf/iclr/NieNSK23}, where a time series is split into a set of $p$-length patches, which constitute the query, key, and value vectors of a self-attention mechanism. We start with a simple case and generalize it to a general situation. In the derivation, we omit the final step of applying softmax to the attention weights, as softmax does not alter the relative order of the attention weights assigned to different time slots in the sequence and will not affect the conclusion.
\\
\textbf{Definition 4.} Given $a,b\in Z, a\neq b$, we set the patch length $p$ to $\text{lcm}(a,b)$, where $\text{lcm}(a,b)$ denotes the least common multiple of $a$ and $b$. It is given a series with compound periodicity $f(t)=c_1 \cos(\omega_1 t)+ c_2 \sin(\omega_2 t)$, where $\omega_1=\frac{2a\pi}{p}$, $\omega_2=\frac{2b\pi}{p}$ and $c_1,c_2 \in R, c_1 > c_2$. There are two subperiod component in $f(t)$: $f_1(t)=c_1\cos(\omega_1 t)$ and $f_2(t)=c_2\sin(\omega_2 t)$. We use $T_1$ and $T_2$ to denote the period length of $f_1$ and $f_2$ respectively.
\\
\textbf{Theorem 1.} 
In $f(t)$, when taking the patch starting from $t_1^{th}$ time slot as the query, the attention weight of the patch starting from $t_2^{th}$ is $\frac{p}{2}[c_1^2 \cos\omega_1\Delta t+c_2^2 \cos\omega_2\Delta t]$, where $\Delta t=(t_2-t_1)$.
\\
\emph{Proof.} Please refer to Appendix \ref{sec:proof1} for more details.

Taking a further look at the attention weight $\frac{p}{2}[c_1^2 \cos\omega_1\Delta t+c_2^2 \cos\omega_2\Delta t]$, it is a linear combination of $\cos\omega_1\Delta t$ and $\cos\omega_2\Delta t$. The first one distributes attention weight according to the periodicity of $f_1$: it assigns the highest attention weight to the time slot that is $nT_1$-slots apart from the query time slot, where $n\in Z$ (i.e. when $\Delta t=nT_1$, $\cos\omega_1\Delta t$ reaches its maximum value). 
Similarly, the second one distributes attention weight according to the periodicity of $f_2$ and assigns the highest attention weight to the time slot that is $nT_2$ apart from the query time slot. 
Moreover, their impact on the attention weight is decided by the amplitudes of their corresponding subperiod. Since $c_1>c_2$, $\cos\omega_1\Delta t$ contributes more to the attention weight. Thus, the periodic information of $f_1$ can obtain higher attention weight and $f_1$ will be reconstructed better. As shown in Fig.~\ref{fig:attention}, the highest attention weights show up at the time slot that $nT_1$-slots apart from the query slot without concerning the subperiod with period length of $T_2$.

To generalize Theorem 1 to a general situation, given a time series $\tilde{f}(t)$ with compound periodicity, we use trigonometric series to decompose it as defined in Definition 5.
\\
\textbf{Definition 5.} Given a compound periodic time series $\tilde{f}(t)$ with period length $p$, we set the patch length to $p$. We decompose $\tilde{f}(t)$ to a linear combination of trigonometric series as: $\tilde{f}(t)=\frac{a_0}{2}+\sum_{n=1}^{\infty}(a_n\cos{\omega_n t}+b_n\sin{\omega_n t})$, where $\omega_n=\frac{2n\pi}{p}$ and $a_n$ and $b_n$ are coefficients for triangonometric series.
\\
\textbf{Theorem 2.}
In $\tilde{f}(t)$, when taking the patch starting from $t_1^{th}$ time slot as the query, the attention weight of the patch starting from $t_2^{th}$ is $\frac{a_0^2p}{4}+\frac{p}{2}\sum_{n=1}^{\infty}(a_n^2+b_n^2)\cos\omega_n\Delta t$, where $\Delta t=(t_2-t_1)$.
\\
\emph{Proof.} Please refer to Appendix \ref{sec:proof2} for more details.

Similar to the analysis of $f(t)$, the attention weight of $\tilde{f}(t)$ is a linear combination of $\cos\omega_n \Delta t$. The subperiods with the higher amplitudes are more decisive to the attention weight distribution and the periodic information of these subperiods can obtain higher attention weights. Thus, the subperiods with higher amplitudes are more likely to reconstruct better, while the subperiods with low amplitudes can be poorly reconstructed.

We show the architecture of each skimming attention layer in Fig.\ref{fig:skimmingAtten}, which aims at preventing the attention mechanism from directly reconstructing time slots by making use of the similarity of adjacent time slots. 
As shown in Fig.\ref{fig:skimmingAtten}, we first use a sliding window with padding to extend the input data $x \in R^{T\times D}$ to $\bar{x} \in R^{T\times (p+1) \times D}$, where $p+1$ denotes the window length of the sliding window. Subsequently, we take each dimension separately (taking the $d^{th}$ dimension as an example) and use the first $p$-length series in each window as the keys and the final time slot in each window as the queries and values. This process is shown in Eq.\ref{eq:extendX}-Eq.\ref{eq:value}, where $[:p+1]$ denotes the time slices from beginning to the $p^{th}$ one. 
\begin{gather}
    \bar{x}=\operatorname{SlidingWindow}(x,p+1,1) \label{eq:extendX}, \\
    q_d=\bar{x}[:,:p+1,d] \label{eq:query}, \\
    k_d=\bar{x}[:,:p+1,d] \label{eq:key}, \\
    v_d=\bar{x}[:,p+1,d] \label{eq:value}.
\end{gather}
Afterward, as shown in Eq.~\ref{eq:attWeight}, we apply a standard attention mechanism to the queries, keys, and values and obtain a set of attention weight $\mathcal{A} \in R^{T\times T}$, where $\mathcal{A}_{i,j}$ denotes the $j^{th}$ attention weight for the $i^{th}$ query. Then, in Eq.~\ref{eq:gateCurve}, we multiply a gate curve $G \in R^{T\times T}$ to the $\mathcal{A}$, where $G[i,j]=1-\exp^{-\frac{(i-j)^2}{\sigma^2}}$, $\sigma$ is a learnable parameter and $*$ denotes an element-wise multiplication. In this way, the attention weights of time slots that are closer to the query are harder to pass through the gate, while the further one can easily get passed. Consequently, we can force the attention mechanism to put more weight on the hopping time slots. Finally, we obtain the reconstruction series in this layer as in Eq.~\ref{eq:output}, where $\tilde{x}_l$ denotes the reconstruction series of the $l^{th}$ skimming attention layer:
\begin{gather}
    \mathcal{A}=  q_d  k_d^T \label{eq:attWeight},\\
    \tilde{\mathcal{A}}= \operatorname{softmax}(\mathcal{A} * G)  \label{eq:gateCurve},\\
    \tilde{x}_l[:,d]=  \tilde{\mathcal{A}}  v_d \label{eq:output}.
\end{gather}

We organize different layers of skimming attention as 
% shown in Eq.~\ref{eq:skimmingLayers}
follows to deal with compound periodic information:
\begin{equation}
    \label{eq:skimmingLayers}
    \begin{split}
        \tilde{x}_l=\operatorname{Skimming}&\operatorname{AttentionLayer}(x_{l}),\\
        x_{l+1}=&x_{l}-\tilde{x}_l,
    \end{split}
\end{equation}
where $x_{l}$ is the $l^{th}$ layer input data and $x_0=x$. There are two benefits to organizing the skimming attention layers in this way. On the one hand, each skimming attention layer skims off the subperiod with the highest amplitude in $x_l$ and the next layer can pay more attention to the subperiod with relatively low amplitude in the remaining series. Consequently, the subperiods with different amplitudes can be reconstructed well. We show the reconstruction series of different Skimming Attention layers in Fig.~\ref{fig:skimSeries}, where it reconstructs subperiods in descending amplitude order. On the other hand, it can also prevent the problem of vanishing gradient like ResNet does, since the input of every layer can be also reduced to $x_{l}=x-\sum_{k=0}^{l-1}\tilde{x}_l$.

\subsection{Neural OT}
\label{sec:NOT}
Besides the periodic patterns, there are still aperiodic but normal fluctuations in task duration time distribution. Since we only pay attention to slow-down anomalies but not others (e.g., the execution speed of homework has significantly increased), we target modeling these non-periodic fluctuations but only hinder the reconstruction of slow-down anomalies.
Inspired by the Optimal Transport (OT) algorithm, we transform a standard OT problem into a neural network and embed it into our model so that our model becomes end-to-end.

We first establish an OT problem and then transform it into a neural network.
For each time slot $t$, we take its reconstruction duration time distribution $\tilde{x}[t] \in R^{1\times d}$ as a source distribution and take its original duration time distribution $x[t] \in R^{1\times d}$ as a target distribution. The OT problem gropes for a transport strategy $P$ to transform the source distribution to the target distribution with a minimum cost $<P*\tilde{x}[t],C>$, where $P[d,s]$ denotes the ratio of $\tilde{x}[t,s]$ transporting to $\tilde{x}[t,d]$, $C[d,s]$ denotes the cost of transporting a unit from $\tilde{x}[t,s]$ to $\tilde{x}[t,d]$ and $*$ denotes element-wise multiplication.  According to the definition of $P$, $P\tilde{x}[t]$ denotes the distribution after applying the transport strategy $P$ to the source distribution $\tilde{x}[t]$, which should approach the target distribution $x[t]$, and the sum of each column of $P$ should be $1$. Thus, we formulate $|P\tilde{x}[t]-x[t]|$ as an optimization goal and the $P^T \vec{1}=\vec{1}$ as a constraint in our OT problem. To reconstruct anomalies except the slow ones, we set $C$ as follows:
% in Eq.~\ref{eq:cost}, 
\begin{equation}
  \label{eq:cost}
  C_{i,j}=
\left\{\begin{matrix}
M[i]-M[j], & i>j
 \\
0, & else,
\end{matrix}\right.
\end{equation}
where $M[i]$ is the midpoint of $i^{th}$ interval in $I$ ($I$ is defined in Definition 3). In this way, only the slow-down distribution shift is penalized by the transporting cost. Based on the setting above, we formulate an OT problem as:
\begin{equation}
  \label{eq:NeuralOT}
  \begin{split}
    \min. \lambda <P*\tilde{x}[t],C> & + \left |P\tilde{x}[t]-x[t] \right |_2,\\
    s.t. P^T \vec{1} & =\vec{1},
  \end{split}
\end{equation}
% in Eq.~\ref{eq:NeuralOT}. 
where $\lambda$ is a hyperparameter belonging to $[0,1]$.

Furthermore, we transform it into a neural network. 
We take $P$ as a trainable parameter. To meet its constraint in the OT problem, we manipulate $P$ as $\operatorname{softmax}(P^T)^T$, and the neural layer is specified as:
% in Eq.~\ref{eq:NOT}. 
\begin{equation}
  \label{eq:NOT}
  \dot{\tilde{x}}=\operatorname{softmax}(P^T)^T \tilde{x}
\end{equation}
Besides, we also introduce the optimization objective of the OT problem to the loss function. 

\subsection{Picky Loss Function}
The reconstruction-based methods assume that there are no anomalies in the training set. However, it is inevitable to have some anomalies in the training set in the scenario of unsupervised learning. Thus, we propose a picky loss function, which adaptively attributes a weight $\mathcal{W} \in R^T$ according to trustworthiness to the loss of each time slot. The more trustful a time slot is, the higher its weight is. 
Inspiring by AnomalyTransformer \cite{DBLP:conf/iclr/XuWWL22}, which points out that the normal points can establish wide-broad informative association along the whole series in attention mechanism whereas the anomalies can only concentrate on adjacent time slots, we utilize the attention weight $\mathcal{A}$ in subsection.~\ref{sec:skimmingAtt} to obtain the weight $\mathcal{W}$. We use a trainable gate curve $\hat{G}\in R^{T\times T}$ to filter out the attention weight in the adjacent part, where $\hat{G}[i,j]=1-\exp^{-\frac{(i-j)^2}{\hat{\sigma}^2}}$ and $\hat{\sigma}$ is a trainable parameter and obtain $\mathcal{W}$ via:
% in Eq.\ref{eq:TrustWeight}.
\begin{equation}
  \label{eq:TrustWeight}
  \mathcal{W}=\operatorname{softmax}[(\mathcal{A}*\hat{G}) \vec{1}].
\end{equation} 
We obtain the final loss function by attributing the weight $\mathcal{W}$ to each time slot and fusing the optimizing objective in Section~\ref{sec:NOT}, resulting in the final loss function as follows: 
\begin{equation}
  \label{eq:loss}
  \mathcal{L}=\sum_{t=0}^T \mathcal{W}[t] (\left | \dot{\tilde{x}}[t]-x[t] \right |_2 +\lambda \left < P * \tilde{x}[t], C \right >).
\end{equation}
% The final loss function is shown in Eq.~\ref{eq:loss}. 
As shown in Fig.~\ref{fig:visualize}, the picky loss function renders lower weights to the anomaly time slots.

\subsection{Anomaly Score}
Since the duration time distribution of different tasks does not distribute uniformly, we split the distribution intervals $I$ according to the distribution density of task duration time. This leads to the heterogeneous importance of the reconstruction errors for different intervals. However, the trivial anomaly score, which adds the reconstruction errors for different intervals together, ignores this heterogeneity. Thus, we use the difference between the task duration time expectations of the original distribution and reconstruction one as the anomaly score:
\begin{equation}
  \label{eq:AnoScore}
  \begin{split}
    \operatorname{AnomalyScore}[t]=\mathbb{E}(\bar{T}(x[t]))-\mathbb{E}(\bar{T}(\dot{\tilde{x}}[t]))\\
    =\sum_{d=0}^{D} x[t,d] * M[d] - \sum_{d=0}^{D} \dot{\tilde{x}}[t,d]  * M[d],
  \end{split}
\end{equation}
% , as shown in Eq.~\ref{eq:AnoScore}, 
where $\operatorname{AnomalyScore}[t]$ denotes the anomaly score of $t^{th}$ time slot, and $\bar{T}(x[t])$ and $\bar{T}(\dot{\tilde{x}}[t])$ denote two variables: the task duration time from distributions $x[t]$ and $\dot{\tilde{x}}[t]$ respectively.

\section{Experiment}
We have made extensive experiments on four datasets to verify the following conclusions:
\begin{itemize}
  \item SORN can achieve the best performances on the four datasets, compared with the SOTA methods.
  \item SORN consumes tolerable time and memory overhead.
  \item SORN is parameter insensitive.
  \item SORN is resistant to noise and lax periodicity.
  \item Each module in SORN has contributed to the performance.
\end{itemize}
\subsection{Experiment Setup}
\textbf{Baseline Methods.} We compare SORN with the SOTA anomaly detection methods: DCdetector \cite{yang2023dcdetector}, TranAD \cite{tuli2022tranad}, AnomalyTransformer \cite{DBLP:conf/iclr/XuWWL22}, VQRAE \cite{kieu2022anomaly}, OmniAnomaly \cite{su2019robust}, MSCRED \cite{zhang2019deep}. Furthermore, we compare SORN with a method specifically designed for slow-down detection: IASO \cite{panda2019iaso} and a method designed for distribution shift detection, feature-shift detection \cite{kulinski2020feature}. 

\textbf{Datasets.} We perform our experiments on four datasets. Two of them (Ali1, Ali2) are monitoring data of industrial cloud clusters from Alibaba. One of them (Mustang) is disclosed by Carnegie Mellon Parallel Data Laboratory, and we label the slow-down anomalies in it manually.  To further verify the impact of different factors on the performance, such as noise, periodicity, slow tasks ratio, and the average task slow-down time, we also introduce a synthetic dataset (Sync) so that we can keep every factor under control. We summarize key statistics of different datasets in Table~\ref{Tab:datasets}. 

\begin{figure*}
  \vspace{-4mm}
  \centering 
  \subfigure[Periodicity strictness of each dataset]{
  \includegraphics[width=0.28\linewidth]{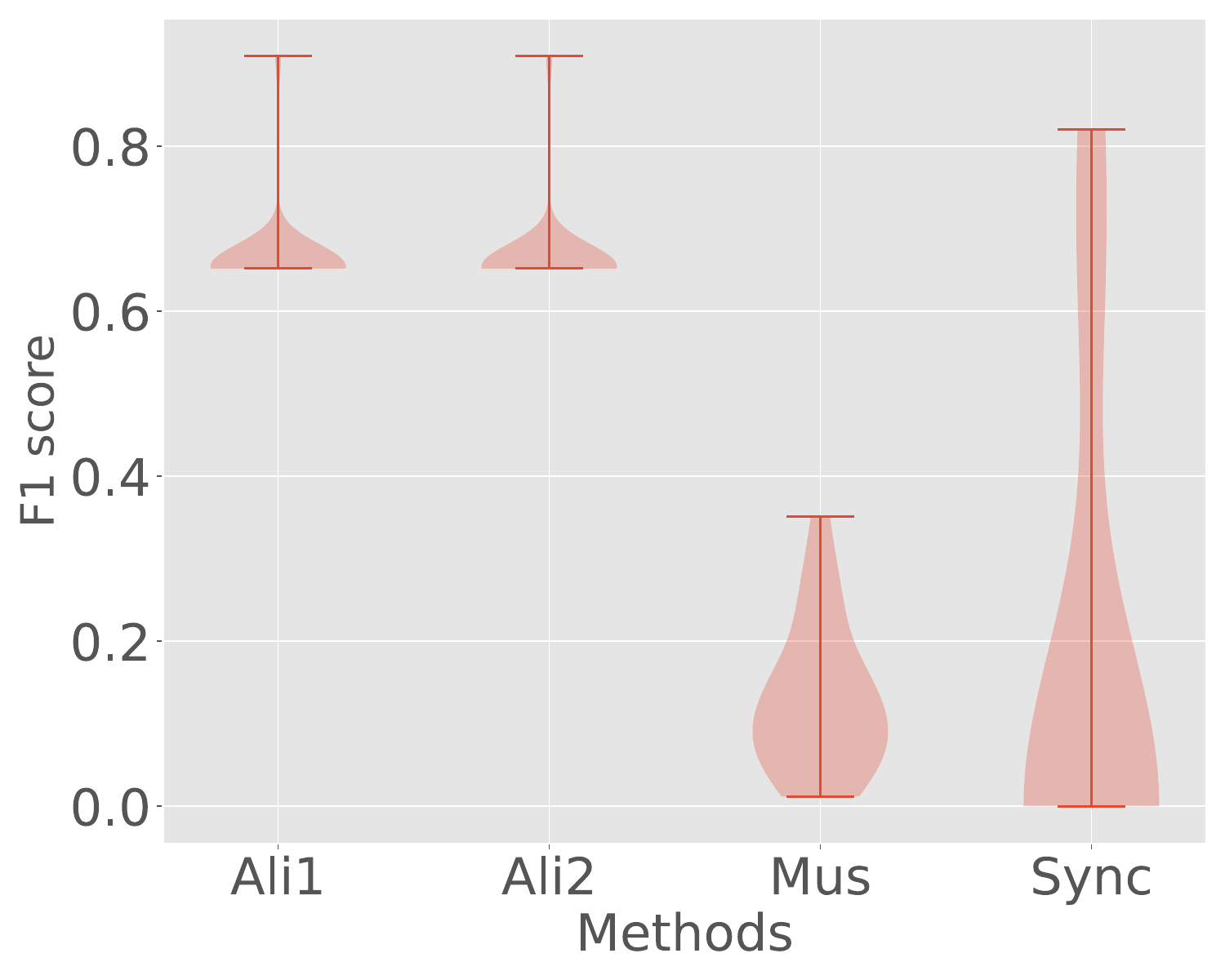}
  \label{fig:periodStrict}
  }
  \hfill
  \subfigure[Time and memory overhead]{
  \includegraphics[width=0.31\linewidth]{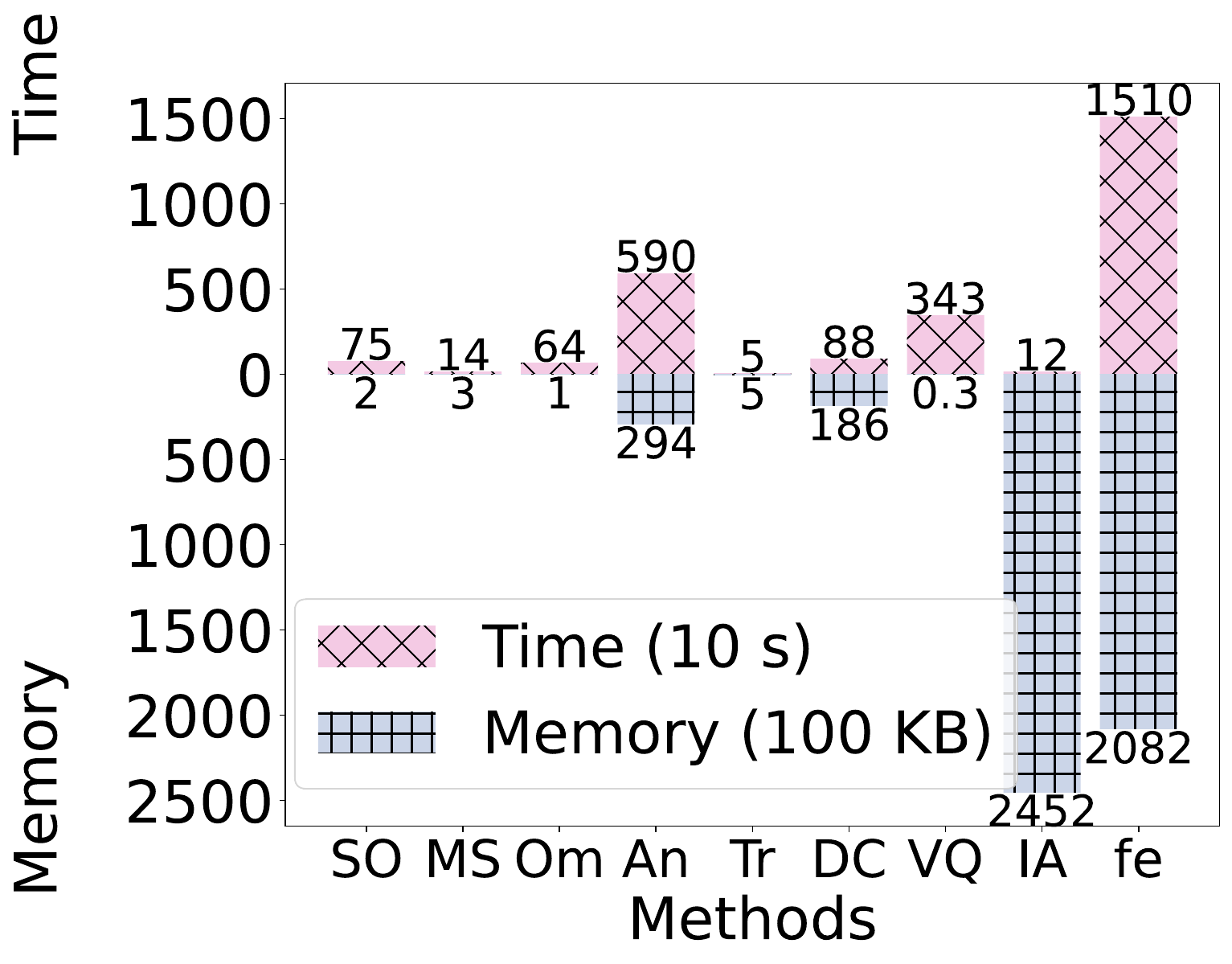}
  \label{fig:overhead}
  }
  \hfill
  \subfigure[Parameter sensitivity]{
  \includegraphics[width=0.35\linewidth]{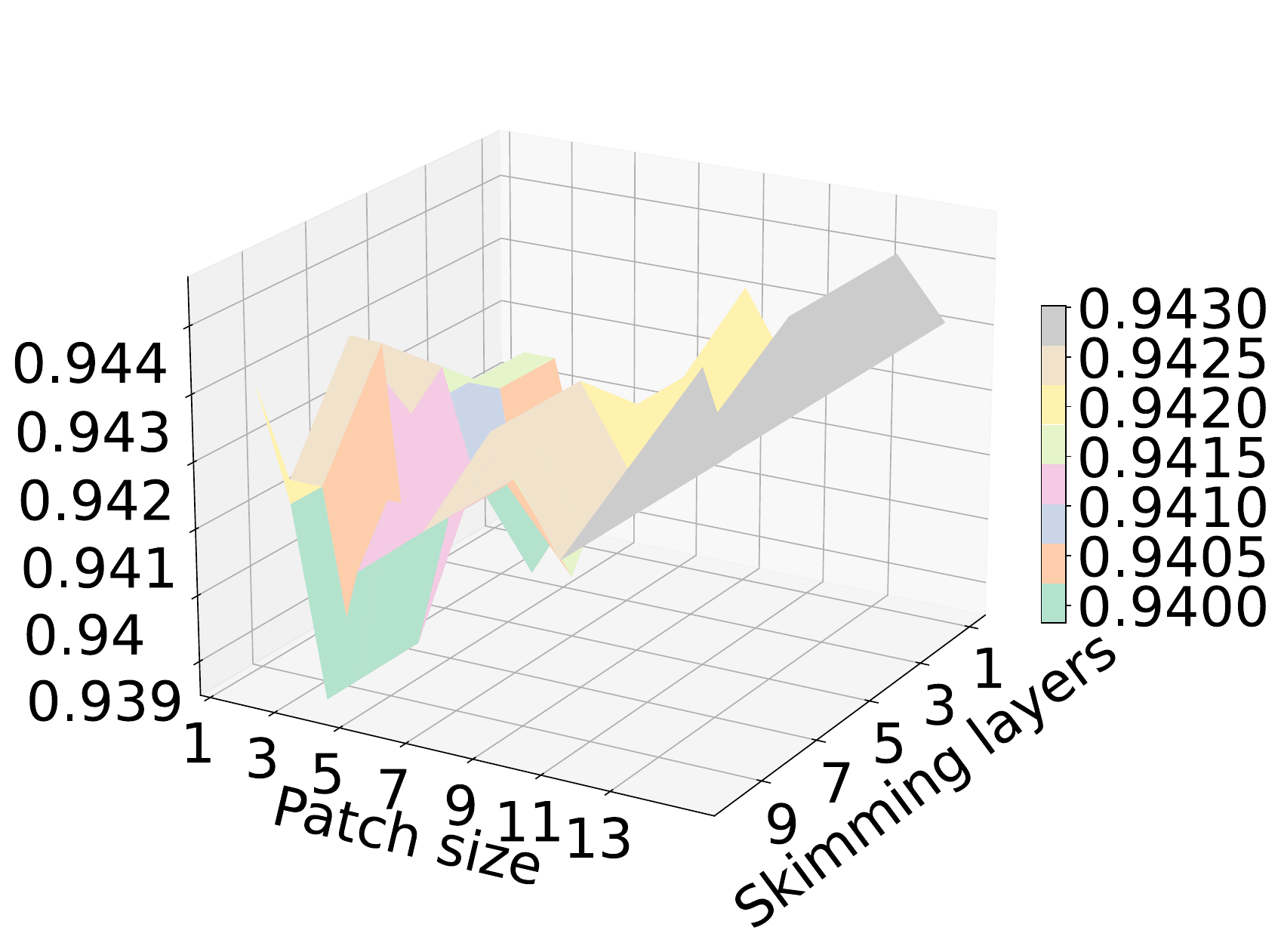}
  \label{fig:sensitivity}
  }\vspace{-4mm}
  \caption{(a) We show the autocorrelation coefficient distribution at the interval of period length for subsets in every dataset. (b) The time and memory overhead of SORN and baselines on Sync dataset. We use the first two characters to stand for each method; (c) The hyperparameter sensitivity of the number of skimming layers and patch size on Sync dataset.}
\end{figure*}

Besides, different datasets exhibit different periodicity strictness. To measure the periodicity strictness of each dataset, for each subset, if it exhibits periodicity we calculate its autocorrelation coefficient at intervals of its period length as its periodicity strictness level, otherwise, we set its periodicity strictness as $0$. We show the periodicity strictness level distribution of subsets in each dataset in Fig.~\ref{fig:periodStrict}. Ali1 and Ali2 show relatively strict periodicity. Mustang shows lax periodicity. Some subsets of Sync show strict periodicity, while others are aperiodic.

For more data preprocessing details, please refer to Appendix.~\ref{sec:data}. 

\begin{table}[]
  \centering
  \renewcommand\arraystretch{1.02}
  \caption{\label{Tab:datasets}Statistics of different datasets.}\vspace{-2mm}
  \begin{tabular}{l|cccc}
  \hline
                    & \multicolumn{1}{l}{Ali1} & \multicolumn{1}{l}{Ali2} & \multicolumn{1}{l}{Mustang} & \multicolumn{1}{l}{Sync} \\ \hline
  Dimension         & 14                       & 14                       & 17                           & 14                       \\
  Anomaly ratio (\%) & 3.71                     & 6.06                     & 7.75                         & 1                        \\ 
  Subsets & 25 & 25 & 1 & 10 \\ \hline
  \end{tabular}%
\end{table}
\begin{table}[]
  \centering
  \renewcommand\arraystretch{1.02}
  \caption{\label{Tab:hyper}The hyperparameters.}\vspace{-2mm}
  \begin{tabular}{lc|lc}
    \hline
    \multicolumn{1}{c}{\textbf{Hyperparameter}} & \textbf{Value} & \multicolumn{1}{c}{\textbf{Hyperparameter}} & \textbf{Value} \\ \hline
    Batch Size                                  & 100            & Learning Rate                               & 0.001          \\
    Skimming Layer of Ali1                      & 10             & Patch Size of Ali1                          & 2              \\
    Skimming Layer of Ali2                      & 6              & Patch Size of Ali2                          & 2              \\
    Skimming Layer of Mut                       & 6              & Patch Size of Mut                           & 2              \\
    Skimming Layer of Sync                      & 6              & Patch Size of Sync                          & 10             \\ \hline
  \end{tabular}
  \end{table}

\textbf{Hyperparameters.} We show some important hyperparameters in Table~\ref{Tab:hyper}, where we use Mut to stand for Mustang.

\begin{figure*}
  \centering 
  \subfigure[The impact of noise and slow task ratio on performance]{
      \includegraphics[width=0.22\linewidth]{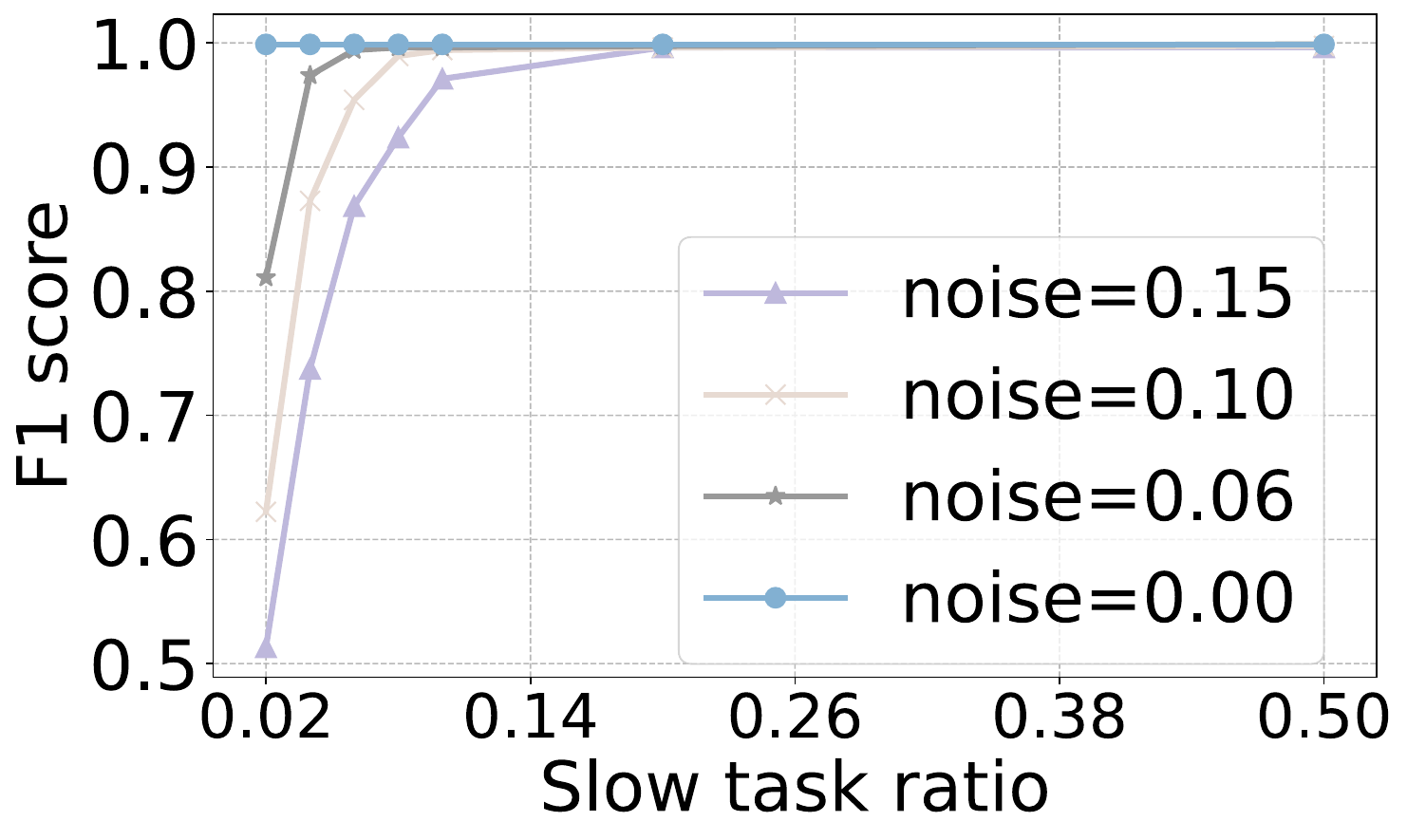}
      \label{fig:slowRatio}
  }
  \hfill
  \subfigure[The impact of periodicity and slow task ratio on performance]{
  \includegraphics[width=0.22\linewidth]{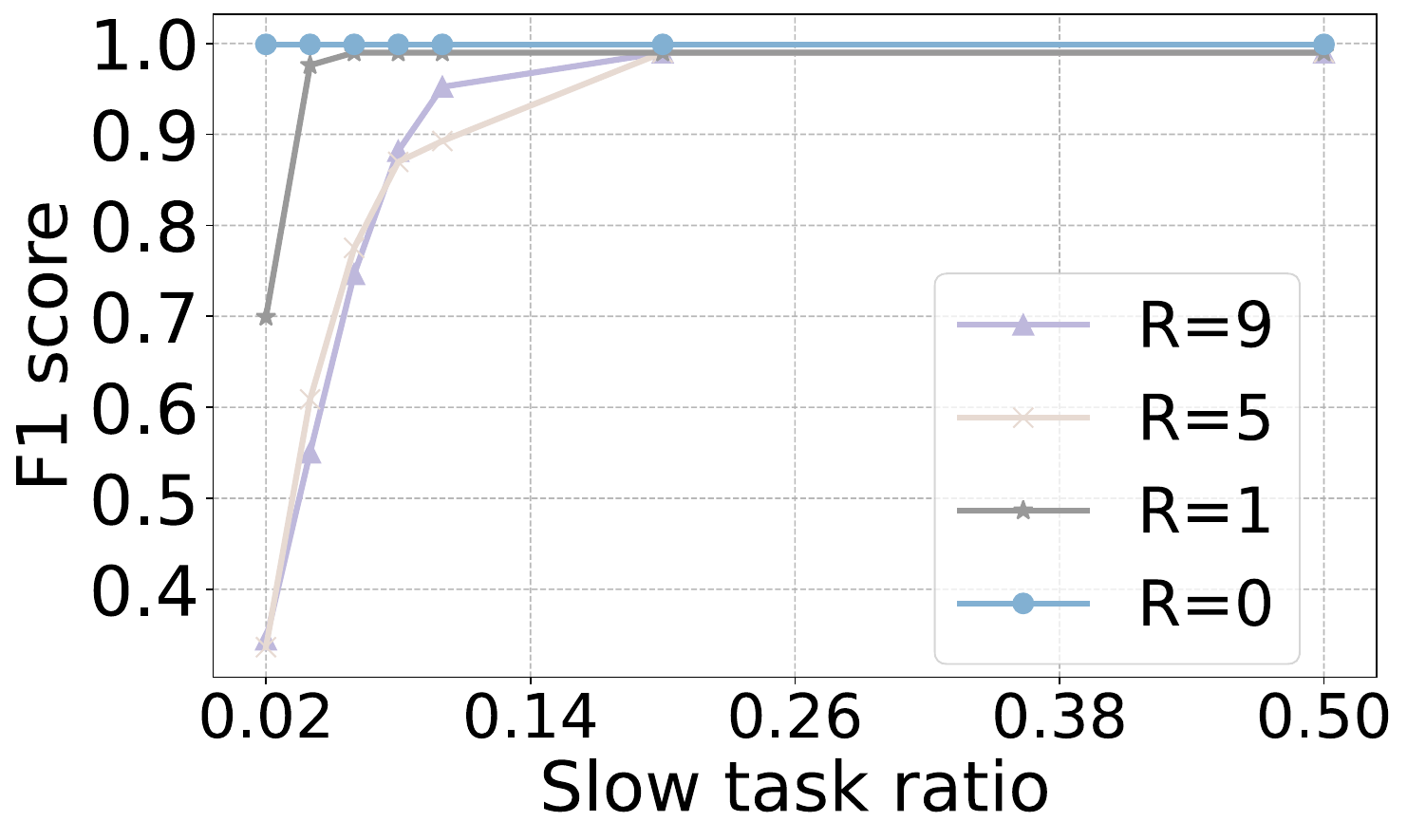}
  \label{fig:ratio2}
  }
  \hfill 
  \subfigure[The impact of noise and average slow-down time on performance]{
  \includegraphics[width=0.22\linewidth]{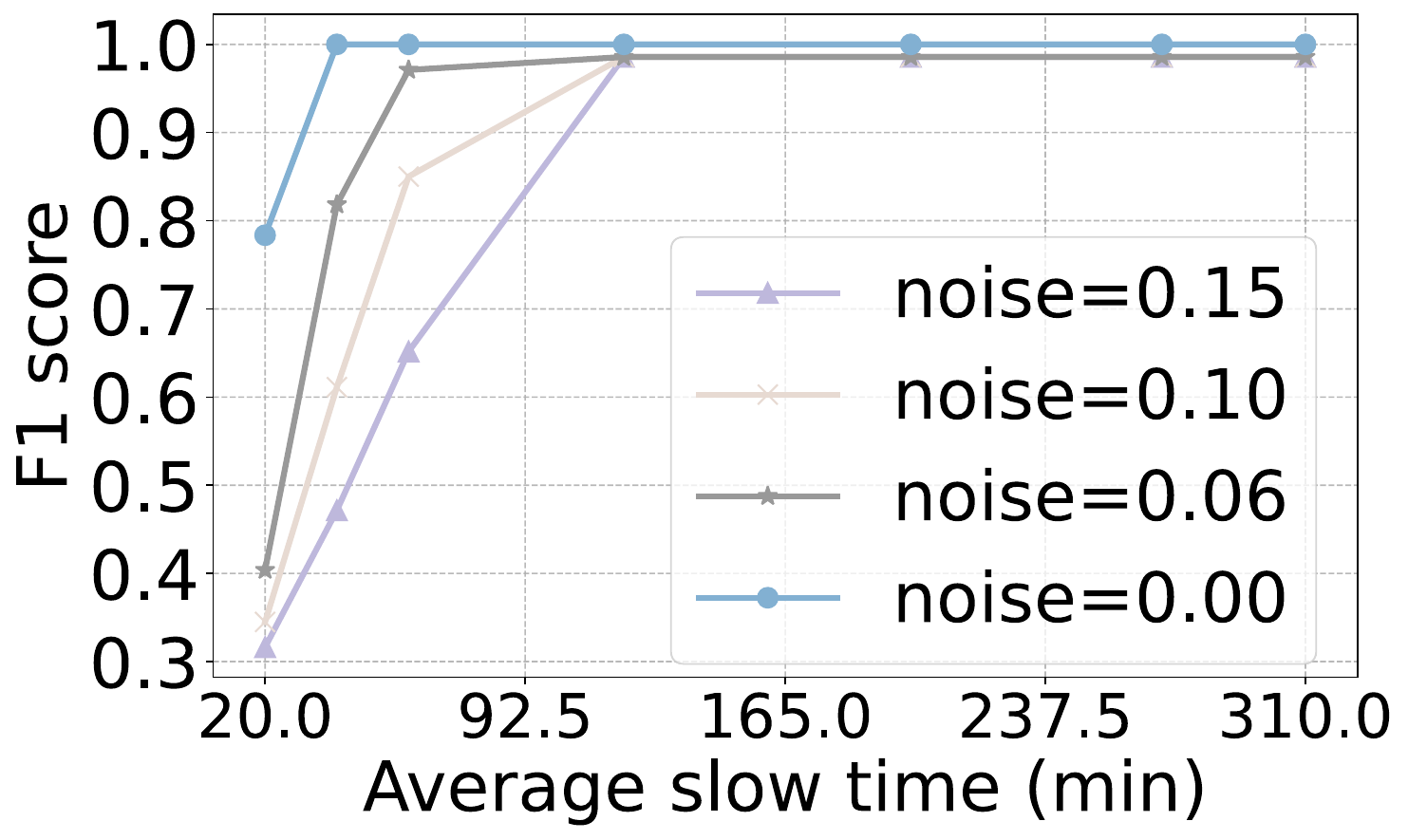}
  \label{fig:slowRange}
  }
  \hfill
  \subfigure[The impact of periodicity and average slow-down time on performance]{
  \includegraphics[width=0.22\linewidth]{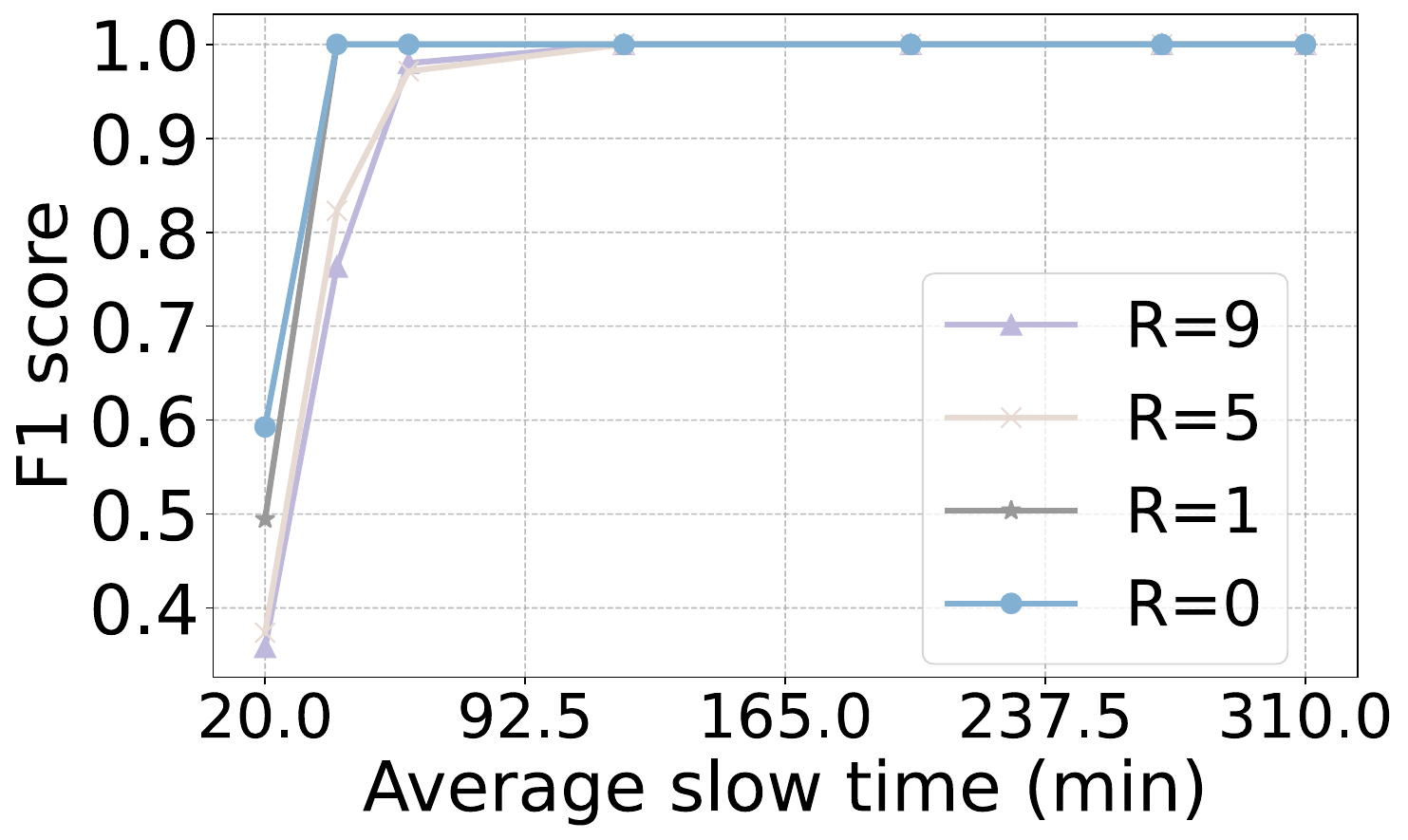}
  \label{fig:slowRange2}
  }\vspace{-4mm}
  \caption{(a) We add noise to the original synthetic time series, whose standard deviation is the maximum amplitude of the original time series multiplied by the "noise" shown in the legend.
  Then, we test the performance of SORN for different slow task ratios. (b) For each period in a periodic time series, we extend it by a scaler which is randomly sampled from $(1,1+R]$. In this way, the original time series will have a lax periodicity. Then, we test the performance of SORN for different slow task ratios. (c) Using the same noise setting as (a), we test the performance of SORN for different average slow-down time. (d) Using the same period setting as (b), we test the performance of SORN for different average slow-down time.}
\vspace{-1em}
\end{figure*}
\textbf{Evaluation Metrics.} We choose three of the most widely-used metrics to measure the performance of our method as many marvelous methods did \cite{li2023prototype, chen2021daemon, chen2022deep}: the precision, recall, and F1 score.

\subsection{Prediction Accuracy}
We take 70\% of each subset as the training set and take the remaining 30\% as the testing set. For each subset, we train a unique model. This training strategy is also adopted by other marvelous works, such as \cite{zhang2019deep,su2019robust,chen2022deep}. We show the performance of SORN and baselines in Tab.~\ref{Tab:perf}, where we use "Pre" and "Rec" to stand for precision and recall respectively, and highlight the best performance as the boldfaced. When SORN achieves the best performance, we underline the best performance among baselines. Otherwise, we underline the second-best performance among all methods. SORN achieves the best F1 scores on all datasets compared with the state-of-the-art methods. Comparing the performance of our method on four datasets, we observe that it performs best on the Ali1 and Ali2 datasets, followed by Mustang and Sync. It can be seen that the effectiveness of our method is positively correlated to the strictness of periodicity in the datasets. It achieves impressive performance on datasets with strict periodicity, while also demonstrating competitive results on datasets with relaxed periodicity or non-periodic characteristics. We will further discuss the impact of periodicity strictness in subsection.~\ref{sec:impact}.
\begin{table*}[]
  \centering
  \renewcommand\arraystretch{1.02}
  \caption{\label{Tab:perf}Average performance of SORN and baselines on subsets of four datasets.}\vspace{-2mm}
  \begin{tabular}{l|lll|lll|lll|lll}
  \hline
                   & \multicolumn{3}{c|}{Ali1}                        & \multicolumn{3}{c|}{Ali2}                        & \multicolumn{3}{c|}{Mustang}                    & \multicolumn{3}{c}{Sync}                         \\ \cline{2-13} 
                   & Pre            & Rec            & F1             & Pre            & Rec            & F1             & Pre            & Rec            & F1             & Pre            & Rec            & F1             \\ \hline
  MSCRED           & 0.841          & 0.981          & 0.878          & 0.928          & 0.988          & 0.954          & 0.871          & 0.960          & 0.896          & 0.717          & 0.874          & 0.779          \\
  Omni             & 0.681          & 0.981          & 0.782          & 0.814          & 0.987          & 0.890          & 0.812          & 0.968          & 0.878          & 0.655          & \textbf{0.997} & 0.787          \\
  AnomalyTr        & \textbf{1.000} & 0.870          & \underline{0.923}          & \textbf{0.999} & 0.763          & 0.857          & \textbf{1.000} & 0.891          & \underline{0.935}    & \textbf{1.000} & 0.680          & \underline{0.809}    \\
  TranAD           & 0.784          & \underline{0.989}    & 0.864          & 0.827          & 0.968          & 0.877          & 0.865          & 0.918          & 0.867          & 0.247          & 0.568          & 0.313          \\
  DCdetector       & \underline{0.984}    & 0.728          & 0.806          & \underline{0.994}    & 0.723          & 0.818          & \underline{0.968}    & 0.718          & 0.799          & 0.936          & 0.406          & 0.567          \\
  VRGAE            & 0.811          & 0.981          & 0.853          & 0.966          & \underline{0.992}    & \underline{0.978}    & 0.871          & 0.959          & 0.905          & 0.624          & 0.794          & 0.648          \\
  IASO             & 0.492          & 0.943          & 0.618          & 0.611          & 0.907          & 0.708          & 0.420          & 0.899          & 0.524          & 0.389          & 0.910          & 0.533          \\
  feature-shift    & 0.533          & \textbf{1.000} & 0.647          & 0.744          & 0.953          & 0.790          & 0.511          & \textbf{1.000} & 0.629          & 0.594          & 0.081          & 0.142          \\ \hline
  SORN$^\dagger$  & 0.891          & \underline{0.989}          & 0.897          & 0.955          & 0.997          & 0.968          & 0.895          & 0.996          & 0.916          & \underline{0.963}    & 0.893          & 0.919          \\
  SORN$^\ddagger$ & 0.944  & 0.969  & 0.939   & 0.960          & 0.967          & 0.955          & 0.912          & 0.971          & 0.919          & 0.939          & 0.832          & 0.874          \\
  SORN$^\S$       & 0.878          & \textbf{1.000}          & 0.891          & 0.950          & 0.997          & 0.965          & 0.925          & \underline{0.996}    & 0.938          & 0.935          & 0.763          & 0.826          \\
  SORN            & \textbf{1.000} & 0.966          & \textbf{0.979}    & 0.980          & \textbf{1.000} & \textbf{0.989} & 0.952          & 0.974          & \textbf{0.958} & 0.956          & \underline{0.926}    & \textbf{0.932} \\ \hline
  \end{tabular}%
  \end{table*}

\subsection{Time and Memory Overhead}
We evaluated both time and memory overhead on a server equipped with a configuration comprising 32 Intel(R) Xeon(R) CPU E5-2620 @ 2.10GHz CPUs and 2 K80 GPUs. We use the checkpoint sizes to stand for the neural network memory overhead and use the time of training model for one epoch to stand for the time overhead. As for the non-neural network methods, IASO and feature-shift detection, we use the maximum memory consumption during its inferring process as its memory overhead. We show the time and memory overhead of different methods in Fig.~\ref{fig:overhead}, where SORN only introduces marginal time and memory overhead compared with some light methods, such as OmniAnomaly, but can achieve better performance on all the datasets. Compared with some transformer-based methods, such as AnomalyTransformer and DCdetector, we use less memory overhead yet achieve better accuracy. In this way, SORN can better meet the real-time requirements of the cloud center.

\subsection{Hyperparameter Sensitivity}
We test the performance of SORN when setting the number of skimming layers and patch size as the Cartesian product of \{1,3,5,7,9\} for skimming layers and \{1,3,5,7,9,11,13\} for patch size. We exhibit the result in Fig.~\ref{fig:sensitivity}. Overall, the performance of SORN is parameter insensitive. As the number of skimming attention layers and patch size increase, the performance of SORN increases in fluctuations.

\subsection{The Impact of Dataset Property}
\label{sec:impact}
We investigate the impact of four factors on the performance of SORN on the Sync dataset: the noise, periodicity strictness, slow task ratio, and average slow-down time in slow-down anomalies. The noise introduced into the Sync data is a random variable with a mean of $0$ and standard deviation of $noise*\mathcal{A}$, where $\mathcal{A}$ is the amplitude of the original time series. 
To manipulate the periodicity strictness, we distort each period of the original series by using a scalar randomly sampled from a distribution $(1,1+R]$ to extend it. When we test the impact of the noise, we make the time series strictly periodic before introducing noise and vice versa.
The results are displayed in Fig.~\ref{fig:slowRatio}-Fig.~\ref{fig:slowRange2}. Generally, when the time series is strictly periodic without any noise, SORN can achieve excellent performance on the Sync dataset. When the noise becomes more variable and the periodicity is more severely distorted, the performance degrades but SORN is still sensitive and accurate: SORN can achieve an F1 score over 0.9 as long as the slow task ratio overpasses 10\% in all conditions of the noise and periodicity strictness explored in our experiment; SORN can achieve an F1 score over 0.9 as long as the average slow-down time overpasses 60 minutes in all conditions of the periodicity strictness and most of conditions of the noise. It is worth noting that 60 minutes is slightly over the maximum interval length in $I$ (50 minutes). Since the maximum interval length is 50 minutes, the slow task with slow-down time less than that may not bring change to $x$. Thus, our model can not distinguish them. If there is a need to improve the sensitivity of SORN to the average slow-down time, we can make it by just substituting the interval division $I$ with a fine-grained one. 

\subsection{Ablation Study}
To evaluate the contribution of each module in SORN, we alternatively remove each submodule and test the performance of the remaining model. Specifically, we denote SORN removing skimming attention as SORN$^\dagger$, denote SORN removing neural OT as SORN$^\ddagger$ and denote SORN replacing picky loss with MSE as SORN$^\S$. When removing the skimming attention mechanism, we replace it with a standard attention. When removing the picky loss, we substitute it with MSE. As shown in Table~\ref{Tab:perf}, the completed SORN achieves the best performance. Thus, each submodule of SORN does contribute to the performance.  

\section{Related Work}
To the best of our knowledge, we are the first to investigate the issue of cluster-wide task slowdowns. While numerous works delve into slow query detection \cite{ma2020diagnosing,zhou2021dbmind} and disk fail-slow detection \cite{lu2023perseus,lu2022nvme}, they primarily focus on detecting slowdowns at the level of individual SQL queries or disks rather than considering the overall aspect. However, detecting slow tasks at the individual level can be unreliable in cloud virtual environments, where task duration time fluctuates randomly and significantly. Single-task slowdowns are common and do not necessarily indicate a cluster malfunction.

Moreover, time series anomaly detection is another relevant area, as we need to capture the normal variation pattern and time dependencies of time series~\cite{zhang2023self,jin2024time}. Time series anomaly detection methods can be broadly categorized into three classes: classical methods \cite{pang2015lesinn,barz2018detecting,nakamura2020merlin,gao2020robusttad}, signal-processing-based methods \cite{zhao2019automatic,alarcon2001anomaly,ma2021jump}, and deep learning-based methods \cite{zhang2022tfad,hundman2018detecting,DBLP:conf/iclr/ZongSMCLCC18, sasal2022w,chen2024lara,sun2023unraveling,xu2024calibrated,xu2023deep}.
Classical methods typically rely on statistical approaches and have relatively low computational overhead. However, they often make specific assumptions that limit their robustness in detecting anomalies in cloud environments \cite{ma2021jump}.
Signal-processing-based methods leverage the sparsity inherent in the frequency domain to reduce computational overhead. However, they may overlook local subtle features \cite{alarcon2001anomaly} or struggle to handle heavy traffic loads in real-time scenarios \cite{ma2021jump}.
Deep learning-based anomaly detection methods have reported promising performance and diversified into various approaches, including prediction-based  \cite{hundman2018detecting, DBLP:conf/iclr/ZongSMCLCC18, sasal2022w,chen2022comprehensive}, reconstruction-based \cite{DBLP:conf/icml/ChenTCDDZ22,you2022unified,jiang2022softpatch,shen2021time,tian2019learning,deng2021graph,ho2023self,chen2024lara}, classification-based \cite{DBLP:conf/iclr/GrathwohlWJD0S20, DBLP:conf/icml/RuffGDSVBMK18, shen2020timeseries,xu2024calibrated,sun2023unraveling}, and perturbation-based methods \cite{cai2022perturbation, stadler2021graph}. Among them, reconstruction-based methods have shown strong advantages over others \cite{kieu2022anomaly}, in which the transformer-based methods have demonstrated good performance recently \cite{DBLP:journals/pvldb/TuliCJ22, DBLP:conf/iclr/XuWWL22, potter2022unsupervised}. However, as we mentioned earlier, the standard attention mechanism may struggle to reconstruct compound periodic time series effectively. 

\section{Conclusion}
In this study, we introduce SORN as a method for detecting cluster-wide task slowdowns in cloud clusters, offering three distinctive features: 1) Skimming Attention, where we provide a theoretical explanation for the limitations of standard attention mechanisms in reconstructing compound periodicity and propose a method to separately reconstruct subperiodic components to ensure accurate reconstruction of both high and low amplitude subperiods; 2) Neural OT, which selectively reconstructs non-slowing exceptional fluctuations; 3) Picky Loss, which assigns weights to time slots in the loss function based on their reliability. Additionally, extensive experiments demonstrate that SORN outperforms state-of-the-art methods in real-world datasets. In the future, we will use large language models for further analysis of the causes of slow-down tasks based on this foundation and employ multi-agent systems for automatic recovery.

\begin{acks}
  This work was supported by the National Science Foundation of China under Grants 62125206 and U20A20173, and in part by Alibaba Group through Alibaba Research Intern Program.
\end{acks}

\newpage
\balance
\bibliographystyle{ACM-Reference-Format}
\bibliography{sample-base}

\appendix
\onecolumn

\section{Proof of Theorem 1}
In the following, we use $\operatorname{AttentionWeight[t_1,t_2]}$ to denote the attention weight of the patch starting from $t_2^{th}$ time slot, when using the patch starting from $t_1^{th}$ time slot as the query. We use the orthogonality of trigonometric functions when deriving Eq.~\ref{eq:att1} to Eq.\ref{eq:att2}. Since $\cos\omega_1 t \cos\omega_1(t+\Delta t)=\frac{1}{2}\cos(\omega_1t+\omega_1(t+\Delta t))+\cos(\omega_1t-\omega_1(t+\Delta t))$, $\sin\omega_2 t\sin\omega_2(t+\Delta t)=-\frac{1}{2}(\cos(\omega_2t+\omega_2(t+\Delta t))-\cos(\omega_2t-\omega_2(t+\Delta t)))$, and $\int_{t_1}^{t_1+p} \cos(2\omega_1t+\omega_1\Delta t)\ dt=0$ (because $p$ is integer multiple of the period length of $\cos(2\omega_1t+\omega_1\Delta t)$), we derive Eq.~\ref{eq:att2} to Eq.~\ref{eq:att3}. Since $\Delta t$ is a constant without relevance to $t$, we derive Eq.~\ref{eq:att3} to Eq.~\ref{eq:att4}.
\label{sec:proof1}
\begin{gather}
  \operatorname{AttentionWeight[t_1,t_2]} =\int_{t_1}^{t_1+p} (c_1 \cos\omega_1 t +c_2 \sin\omega_2 t) [c_1 \cos\omega_1(t+\Delta t)+c_2\sin\omega_2(t+\Delta t)]\ dt \label{eq:att1} \\
  \hspace{2.2cm} =\int_{t_1}^{t_1+p} c_1^2 \cos(\omega_1 t) \cos\omega_1(t+\Delta t)+c_2^2 \sin(\omega_2 t) \sin\omega_2(t+\Delta t) \ dt \label{eq:att2} \\
  \hspace{-0.6cm} =\int_{t_1}^{t_1+p} \frac{1}{2}c_1^2 \cos(\omega_1\Delta t) +\frac{1}{2}c_2^2 \cos(\omega_2\Delta t)\ dt \label{eq:att3} \\
  \hspace{-2.3cm} = \frac{p}{2} (c_1^2 \cos\omega_1\Delta t+ c_2^2 \cos\omega_2 \Delta t) \label{eq:att4}
\end{gather}

\section{Proof of Theorem 2}
\label{sec:proof2}
We prove Theorem 2 in a similar way as in Theorem 1.
\begin{equation}
  \begin{split}
  \operatorname{AttentionWeight}[t_1,t_2] &=
  \int_{t_1}^{t_1+p} (\frac{a_0}{2}+ \sum_{n=0}^{\infty} a_n \cos\omega_n t +b_n \sin\omega_n t) \cdot 
  [\frac{a_0}{2}+ \sum_{n=0}^{\infty} a_n \cos\omega_n (t+\Delta t) +b_n \sin\omega_n (t+\Delta t)]\ dt \\
  &= \frac{a_0^2p}{4}+ \sum_{n=0}^{\infty}  \int_{t_1}^{t_1+p} a_n^2 \cos\omega_nt \cos\omega_n (t+\Delta t) +
   b_n^2\sin\omega_nt \sin\omega_n(t+\Delta t)\ dt \\ 
  &= \frac{a_0^2p}{4}+ \frac{p}{2} \sum_{n=0}^{\infty} (a_n^2+b_n^2) \cos \omega_n \Delta t
  \end{split}
\end{equation}

\section{Data preprocessing}
\label{sec:data}
The code and some datasets are available at https://github.com/gyhswtxnc/SORN.
\begin{itemize}[leftmargin=*]
  \item \textbf{Ali1 \& Ali2} (periodic): We collect these datasets by tracing 25 industrial cloud clusters from Alibaba for 15 days. Most of the labels in these two datasets are assigned manually according to the experience of our engineers. Some of the labels are assigned according to our customer's feedback. These two datasets were collected on server clusters in different regions, and there is a significant difference in the anomaly proportion between them. Each subset in Ali1 and Ali2 stands for a cluster.
  \item \textbf{Mustang} (lax periodic) \cite{amvrosiadis2018diversity}: Mustang is a dataset that records task duration time for 5 years. We preprocess the original dataset as shown in Appendix.~\ref{sec:data} and label the slow-down anomalies manually. Then, we equally divide the five years of tracing data into 35 intervals and constitute 35 subsets.
  \item \textbf{Sync} (mixture of periodic and aperiodic): We synthesize this dataset by combining cosine waves with different frequencies and amplitudes. Then, we manually insert noise, distorted period and slow-down anomalies. 
\end{itemize}
For every dataset, we count a task duration time distribution $I$ at each time slot and divide the intervals in $I$ according to the distribution density of the execution time. We show the interval division for every dataset in Tab.~\ref{Tab:I}.
\begin{table*}[h]
  \centering
  \caption{\label{Tab:I}The interval division for each dataset.}
  \begin{tabular}{c|l}
  \hline
  \textbf{Dataset} & \multicolumn{1}{c}{\textbf{Edges of $I$}}                                             \\ \hline
  Ali1             & \{0, 10, 20, 30, 40, 70, 110, 150, 190, 230, 280, 330, 380, 430\}                     \\
  Ali2             & \{0, 10, 20, 30, 40, 70, 110, 150, 190, 230, 280, 330, 380, 430\}                     \\
  Mut              & \{0, 5, 10, 20, 30, 40, 70, 110, 150, 190, 230, 280, 330, 380, 430, 900, 1200, 9000\} \\
  Sync             & \{0, 10, 20, 30, 40, 70, 110, 150, 190, 230, 280, 330, 380, 430\}                     \\ \hline
  \end{tabular}%
  \end{table*}

\section{Hyperparameter searching space}
We use grid-search to figure out the optimal hyperparameter settings. We list the ranges for important hyperparameters in Tab.\ref{Tab:hyperRange}.
\begin{table*}[]
  \centering
  \caption{\label{Tab:hyperRange}The searching ranges for important hyperparameters.}
  \begin{tabular}{l|l}
  \hline
  \multicolumn{1}{c|}{\textbf{Hyperparameter}} & \multicolumn{1}{c}{\textbf{Searching Range}} \\ \hline
  Skimming layers                              & \{1,3,5,7\}                                    \\
  Patch size                                   & \{2,3,4,5,7,9,11,15\}                          \\
  Window length                                & \{10,20,30,40,50,80\}                          \\
  Learning rate                                & \{0.0001,0.001,0.01\}                          \\ \hline
  \end{tabular}%
\end{table*}
\section{Baselines introduction}
\begin{itemize}[leftmargin=*]
  \item \textbf{DCdetector}: DCdetector is one of the most SOTA anomaly detection methods, which assembles a novel dual attention asymmetric design and a pure contrastive loss.
  \item \textbf{TranAD}: TranAD is an influential and novel anomaly detection method, which is assisted by meta-learning and shows the high accuracy of anomaly detection.
  \item \textbf{AnomalyTransformer}: AnomalyTransformer is one of the founders who introduced the deep transformer into the area of anomaly detection, which is verified with strong performance.
  \item \textbf{VQRAE}: VQRAE is a novel and sharp anomaly detection method, which also delves into the problem that there are anomalies in the training set. Thus, we also include this method in our baseline.
  \item \textbf{OmniAnomaly}: OmniAnomaly is one of the most widely-recognized and widely-used anomaly detection methods with small time and memory overhead.
  \item \textbf{MSCRED}: MSCRED is an anomaly detection method garnering widespread attention with strong efficacy, which not only considers the temporal correlation but also takes the interdependency between features into account.
  \item \textbf{IASO}: IASO is a method specifically designed to detect the slow-down data retrieval of disks.
  \item \textbf{Feature-shift detection}: The feature-shift detection method is designed to detect whether the distribution of features has shifted.
\end{itemize}
\end{document}